%% file: main.tex
\documentclass[a4paper]{article}

\usepackage[english]{babel}
\usepackage[utf8x]{inputenc}
\usepackage[T1]{fontenc}  
 
\usepackage[a4paper,top=3cm,bottom=2cm,left=3cm,right=3cm,marginparwidth=1.75cm]{geometry}

\usepackage{amsmath}
\usepackage{graphicx}
\usepackage[colorinlistoftodos]{todonotes}
\usepackage[colorlinks=true, allcolors=blue]{hyperref}
\usepackage{authblk}

\usepackage{fullpage}
 
\input{newcommand}

\title{Online Learning for Non-Stationary A/B Tests}

\author[1]{Andr\'es Mu\~noz Medina}
\author[1]{Sergei Vassilvitskii}
\author[2]{Dong Yin}

\affil[1]{Google AI, New York}
\affil[2]{Department of EECS, UC Berkeley}

\begin{document}
\maketitle
 
\begin{abstract}
The rollout of new versions of a feature in modern applications is 
a manual multi-stage process, as the feature is released to ever larger
groups of users, while its performance is carefully monitored. This kind of A/B 
testing is ubiquitous, but suboptimal, as the monitoring requires heavy
human intervention, is not guaranteed to capture consistent, but short-term
fluctuations in performance, and is inefficient, as better versions take a long 
time to reach the full population. 

In this work we formulate this question as that of expert learning, and 
give a new algorithm {\em Follow-The-Best-Interval}, FTBI,  that works 
in dynamic, non-stationary environments. Our approach is practical, simple, and efficient, and has rigorous guarantees on its performance. Finally, we perform a thorough evaluation on synthetic and real world datasets and show that our approach outperforms current state-of-the-art methods.
\end{abstract}

\input{paperbody}
\bibliographystyle{ieeetr}
\bibliography{ref}
\input{proofs}

\end{document}

%% file: newcommand.tex
\usepackage{amsmath}
\usepackage{amsfonts}
\usepackage{graphicx}
\usepackage{enumerate} 

\usepackage{dsfont}
\usepackage{algorithm}
\usepackage{algorithmic}

\usepackage{hyperref}
\hypersetup{
  colorlinks=true,
  linkcolor={blue}
}
\newcommand{\D}{\mathcal{D}}
\newcommand{\cA}{\mathcal{A}}
\newcommand{\cF}{\mathcal{F}}
\newcommand{\cU}{\mathcal{U}}
\newcommand{\cV}{\mathcal{V}}
\newcommand{\cW}{\mathcal{W}}

\newcommand{\cref}[1]{Claim~\ref{#1}}

\newcommand{\ind}{\mathds{1}}
\DeclareMathOperator{\E}{\mathbb{E}}

\newcommand{\ignore}[1]{}

\newtheorem{theorem}{\bf Theorem }
\newtheorem{definition}{\bf Definition}

\newtheorem{lemma}{\bf Lemma}

\newcommand{\mat}[1]{\boldsymbol{#1}}
\renewcommand{\r}{\mat{r}}
\newcommand{\T}{\mathcal{T}}
\newcommand{\I}{\mathcal{I}}

\newcommand{\Long}{\mathcal{L}}
\newcommand{\N}{\mathbb{N}}
\newcommand{\ACT}{\text{ACTIVE}}
\DeclareMathOperator*{\argmax}{\arg\!\max}

\newcommand{\prob}[1]{\mathbb{P}\left\{#1\right\}}
\newcommand{\probs}[1]{\mathbb{P}\{#1\}}
\newcommand{\EXP}[1]{\mathbb{E}\left[#1\right]}
\newcommand{\EXPS}[1]{\mathbb{E}[#1]}

\newcommand{\abss}[1]{|#1|}

%% file: paperbody.tex
\section{Introduction}
\label{sec:intro}
Whether it is a minor tweak, or a major new update, releasing a new version of a running system is a   stressful time. While the release has typically gone through rounds of offline testing, real world testing often uncovers additional corner cases that may manifest themselves as bugs, inefficiencies, or overall poor performance. This is especially the case in machine learning applications, where models are typically trained to maximize a proxy objective, and a model that performs better on offline metrics is not guaranteed to work well in practice. 

The usual approach in such scenarios is to evaluate the new system through a series of closely monitored A/B tests. The new version is usually released to a small number of customers, and, if no concerns are found and metrics look good, the portion of traffic served by the new system is slowly increased. 

While A/B tests provide a sense of safety in that a detrimental change will be quickly observed and corrected (or rolled back), they are not a silver bullet. First, A/B tests are labor intensive---they are typically monitored manually, with an engineer, or a technician, checking the results of the test on a regular basis (for example, daily or weekly).  Second, the evaluation is usually dependent on average metrics---e.g. increasing clickthrough rate, decreasing latency, etc. However, fluctuations in that metric can be easily missed---for instance a system that performs well during the day, but lags at night, may appear better on a day-to-day view, and thus its sub-optimality at night is missed.  Finally, due to their staged rollouts, A/B tests are inefficient---the better version of the system remains active only on a small subset of traffic until a human intervenes, even if it is universally better right off the bat. 

We propose a method for automated version selection that addresses all of the inefficiencies described above. At a high level, we cede the decision of whether to use the old version or a new version to an automated agent. The agent repeatedly evaluates the performance of both versions and selects the better one automatically. As we describe below, this, by itself, is not a new approach, and is captured in the context of {\em expert} or {\em bandit learning} in the machine learning literature. However, the vast majority of previous work optimizes for the average case, where the distribution of real world example is fixed (but unknown). This modeling choice cannot capture short term but consistent fluctuations in efficacy of one system over another.  The main contribution of this work is a simple, practical, and efficient algorithm that selects the best version even under non-stationary assumptions; and comes with strong theoretical guarantees on its performance. 

The paper is organized as follows: first we describe the problem of learning with expert advice and discuss previous work related to the contents of this paper. In Section~\ref{sec:setup} we formalize the expert learning scenario we will be analyzing. Our proposed algorithm and its performance analysis are presented in Sections~\ref{sec:algorithm} and~\ref{sec:theory} respectively. Finally, in Section~\ref{sec:experiments} we conduct extensive empirical evaluations demonstrating the advantages of our algorithm over the current state of the art.

\subsection{Expert Learning}
Learning with expert advice is a classical setting in online learning. Over a course of $T$ time steps, a player is repeatedly asked to decide on a best action in a particular setting. For example, in parallel computing, the agent may need to decide how many workers to allocate for a particular computation; in advertising auctions, the auctioneer needs to decide on a minimum (reserve) price to charge; in ranking systems, the algorithm needs to decide which answer to rank at the top, and so on.

The player does not make these decisions in a vacuum, instead he has access to a number of {\em experts}. Each of the experts may provide her own answer to the query; in our setting each expert represents a different version of the system. At a high level, the goal of the player is to perform (almost) as well as the best expert overall. The difference between the performance of the player and the best expert is known as {\em regret}, thus the player's goal is equivalent to minimizing his regret. Note that if all experts perform poorly, the player is not penalized for not doing well, whereas if one expert does well, the player is best served by honing in on this `true'  expert and following her recommendations. 

To prove bounds on the worst case regret achieved by the player following a particular strategy, we must make assumptions on the distribution from which queries are drawn. In the most studied versions of the problem, the environment is either stochastic (queries are drawn from a fixed, but unknown, distribution), or adversarial (where no assumptions are made). While it is not realistic to assume a fully stochastic setting, neither is an adversarial setting a good model for distributions observed in practice. Rather, the environment is better represented as a series of stochastic intervals, that may have sharp and unpredictable transitions between them. For a concrete example, consider the problem of predicting time to travel from point A to point B. If we have two experts, one performing well when the roads are congested, and another when the roads are relatively free, then a day may have multiple periods that are locally stochastic, but are drastically different from each other. Moreover, while some congested times are predictable, such as the morning and evening rush hour, those caused by accidents, special events, and holidays are not, and represent the scenario we target. 

To model this setting, several notions such as {\em shifting regret}~\cite{herbster1998tracking} and {\em adaptive regret}~\cite{hazan2007adaptive} have been introduced, and these works try to ensure that the performance of the algorithm over sub-intervals of time is comparable to the best expert for that time interval. Several algorithms were developed for this setting, and we provide a more comprehensive review of these algorithms in Section~\ref{sec:previous_work}. Among these algorithms, AdaNormalHedge developed by Luo and Schapire~\cite{luo2015achieving} represents the current state of the art, since it is relatively fast, parameter-free, and gives provable regret bounds. 

While AdaNormalHedge seems like the answer we seek, it has never been empirically validated and has some disadvantages which render it less effective in practice. The straightforward implementation has update time linear in the time elapsed since the start, and, while compression tricks borrowed from streaming literature reduce the bound to polylogarithmic in theory, these updates are nevertheless slow and costly in practice. However, the main drawback of this algorithm is it is not intuitive, and thus is hard to debug and interpret. 

In this work, we propose a different approach. Our algorithm, {\em Follow-The-Best-Interval} (FTBI) builds on the standard Follow-The-Leader (FTL) algorithm for purely stochastic environments, and is easy to explain. We prove (see Theorem~\ref{thm:ftbi}) that in the case of two experts, the algorithm has regret polylogarithmic in $T$, and quadratic in the number of changes in the environment. We  also show that the update time complexity of our algorithm is $O(\log T)$, which is provably better than that of AdaptiveNormalHedge even with the streaming technique. The regret bound of our algorithm is slightly weaker than that given by AdaNormalHedge; however as we show through extensive simulation and real-data experimental results, in practice FTBI performs significantly better. 

\subsection{Related work}\label{sec:previous_work}
Our setup best matches that of learning with expert advice. 
This area of machine learning has been
well studied since the pioneering works of
\cite{cesa1997use, freund1995desicion, littlestone1994weighted,
vovk1995game}.  There are two common assumptions on the nature of the
rewards observed by the agent: stochastic and
adversarial rewards.  In the stochastic setting, Follow-The-Leader
(FTL) is an easy and intuitive algorithm that can achieve
constant regret; while for the adversarial setting, the Weighted
Majority algorithm~\cite{freund1995desicion, littlestone1994weighted}
and the Follow the Perturbed Leader
algorithm~\cite{hutter2005adaptive, kalai2005efficient} are the most
commonly used algorithms with regret in $\tilde{O}(\sqrt{T})$.

The aforementioned algorithms measure regret against a single fixed
best expert.  This is a relatively simple and well-understood
scenario. By contrast the non-stationary environments generally pose greater 
challenges in the design of regret minimizing algorithms. In this setting,
the full learning period $[T]$ is usually partitioned into a few
segments; and the algorithm aims to compete against the sequence of
best experts over each segment. The regret in this setting is also
known as \emph{shifting regret} (See Section \ref{sec:setup} for a precise definition). 
Similar to the non-adaptive scenario,
we model the input as coming from stochastic or adversarial settings.  In the
stochastic setting, the rewards of each expert remain i.i.d. in each
segment, but the distribution may change across the segments. For the
adversarial setting, no assumptions are made as to how the rewards are
generated.  The first analysis of online learning with shifting
environments was given by Herbster and
Warmuth~\cite{herbster1998tracking}. The authors proposed the
Fixed-Share algorithm (and a few variants) to achieve low shifting
regret in the adversarial setting.  The idea of Fixed-Share, and other
derivations of this
algorithm~\cite{adamskiy2012closer,adamskiy2012putting,bousquet2002tracking,
cesa2012mirror,herbster2001tracking}, is to combine the Weighted
Majority algorithm with a mixing strategy.  In general these
algorithms are guaranteed to achieve regret in $\tilde{O}(\sqrt{T})$
with an additional penalty for shifting.

A different framework for solving this problem is the so called
sleeping expert technique, originally introduced by Freund et
al.~\cite{freund1997using}. Although the algorithm achieves good shifting regret
bounds, in its original form, this algorithm has prohibitively high computational costs. To address this issue, Hazan
and Seshadhri~\cite{hazan2007adaptive} propose the Follow the Leading
History (FLH) algorithm with a data streaming technique to reduce the
time complexity of sleeping expert algorithms. The authors provide an
algorithm with logarithmic shifting regret. However, this bound holds
only for rewards distributed according to an exp-concave distribution. Since
we make no assumption on the distribution generating our rewards, we cannot
extend the results of FLH to our setup.

One final line of work dealing with shifting regret under adversarial
rewards is the so called strongly adaptive online learning
(SAOL)~\cite{daniely2015strongly, jun2016improved, zhang2017strongly}.
SAOL framework aims to achieve low regret over any
interval, and thus is strictly harder than achieving low shifting regret.
Despite this difficulty, there are algorithms that achieve regret in $O(\sqrt{T})$, matching the traditional shifting regret scenario. Moreover,  without any assumptions on the 
mechanism generating the rewards, this regret is in fact 
tight. In practice however, rewards are hardly fully
adversarial. While we certainly don't expect rewards to be i.i.d. over
the full time period, it is generally reasonable to assume we can
model rewards as a stochastic process with shifts. For instance,
rewards can be i.i.d. throughout a day but change on weekends or
holidays. When this is the case,
one can obtain exponential improvements and achieve regret of
$O(\text{poly}(\log T))$.  The design of an algorithm with this regret bound
in the stochastic setting with shifts was proposed as an
open problem by Warmuth et al.~\cite{warmuth2014open}. A few solutions have been
given since then. Sani et al. ~\cite{sani2014exploiting} propose an
algorithm that can achieve low shifting regret in both stochastic and
adversarial settings, however, their algorithm is not parameter-free and
it requires proper tuning and  knowledge of the number of
shifts. Luo and Schapire ~\cite{luo2015achieving} introduced the
AdaNormalHedge (ANH) algorithm, a parameter free approach that achieves 
shifting regret of $O(\frac{N}{\Delta} \log T)$, where $N$ is the number of shifts
and $\Delta$ captures the gap between the best and second best choices. 

While this bound is strictly better than the regret bound we provide for FTBI, we will extensively demonstrate that in practice FTBI consistently outperforms ANH both in speed and accuracy. Moreover, FTBI is a much more intuitive and explainable algorithm since it is a simple generalization of the classic Follow The Leader approach (FTL). By contrast, the state of ANH depends on an streaming technique \cite{hazan2007adaptive,luo2015achieving} which offers no transparency in the decisions made by the algorithm.

\section{Setup}
\label{sec:setup}
Let $T > 0$ denote a time horizon and $[T] = \{1, \ldots, T\}$. We
consider the problem of designing an automated agent that, at every
time $t \in [T]$ chooses between two version of a system $x_t \in
\{V_1, V_2\}$. (The setting easily extends to $K$ versions,  and we will 
explore the performance of our algorithm for $K > 2$ in Section \ref{sec:experiments}.) 
After the choice is made, a reward vector 
$\r_t = (r_t(V_1), r_t(V_2))\in [0,1]^2$ is revealed and the agent obtains a reward
$r_t(x_t)$. We consider a full information, as opposed to a bandit setup since in practice we can run simultaneous experiments that can provide us with the reward information for each version. 

The goal of the monitoring system is to maximize the expected cumulative reward $\EXPS{\sum_{t=1}^T r_t(x_t)}$. For instance,
$V_1$ and $V_2$ could be two versions of a network allocation
algorithm and $\r_t$ corresponds to the average number of queries per
second the network can handle. We will abuse our notation and denote by $r_t(k) 
= r_t(V_k)$ the reward of using version $V_k$.

 We now describe the
reward generation mechanism. Let $0 =
\tau_0 < \tau_1 \ldots < \tau_N = T$ be a partition of $[T]$, where
$N$ denotes the total number of shifts.  For every $i\in[N]$, let
$\T_i = [\tau_{i-1} +1, \tau_i]$ be the $i$-th \emph{segment} of the
partition. We assume that for $t \in \T_i$, reward vectors $\r_t$
are drawn i.i.d according to an unknown distribution $\D_i$. We make no
assumptions on $\D_i$ in particular $\D_i$ can differ from $\D_j$
for $i \neq j$.

Notice that if we consider each version to be an expert, we
can cast our problem as that of learning with expert advice under
\emph{stochastic rewards} and \emph{changing environments}. Therefore, borrowing from the expert learning literature, we measure the performance of our agent using the shifting pseudo-regret.  

\begin{definition}
 For every $i \in [N]$, $t \in T_i$ and $k
\in \{1,2\}$, let $\mu_i(k) 
 =\E[r_t(k)]$  denote the expected reward of version $V_k$ with
 respect to the  distribution $\D_i$.  Let  $k_i^* = \argmax_{k}
 \mu_i(k) $ and $\mu^*_i = \mu_i(k_i^*)$ denote the optimal version
 over segment $\T_i$ and its reward respectively. The  shifting pseudo-regret is
 defined  by 
 \begin{equation*}
R_T := \sum_{i=1}^N \sum_{t = \tau_{i-1}+1}^{\tau_i} \mu_i^*  -
\EXPS{r_t(x_t)}, 
\end{equation*}
where expectation is taken over both the reward vector and the
randomization of the monitoring algorithm.
\end{definition}

 A successful algorithm for this scenario is one for which $R_T$ is in $o(T)$. That is, the agent \emph{learns} to choose the optimal version in a sublinear number of rounds.

 Let $\Delta_i= \min_{k \neq   k_i^*} \mu_i^* - \mu_i(k)$  denote the
 expected reward gap between the best and next best version in
$\T_i$, and let $\Delta = \min_{i \in [N]} \Delta_i$. Throughout the
paper we will assume that $\Delta > 0$. Notice that this is without
loss of generality as the case $\Delta = 0$ is uninteresting since
playing either version would have yielded the same expected reward.

\section{Follow-The-Best-Interval}
\label{sec:algorithm}
We begin this section by describing the Follow-The-Leader (FTL)
algorithm, a simple yet powerful algorithm for learning with expert
advice under stochastic rewards (with no shifts). FTL maintains a
weight $W_t(k) = \sum_{s=1}^{t} r_s(k)$ for each expert $k \in
\{1,2\}$. Weights correspond to the cumulative reward seen thus
far. At time $t$, FTL chooses the expert with the highest weight, in
other words, the leader. It is well known that if rewards are sampled
i.i.d., FTL is guaranteed to achieve constant pseudo-regret in
$O(\frac{1}{\Delta})$.

If we had access to the switching times $\tau_1$, $\ldots$,
$\tau_N$, we could simply run FTL and
restart the weights kept by the algorithm at every breakpoint
$\tau_j$. This would guarantee pseudo-regret of $O\big(\frac{N}{\Delta}\big)$.
In the absence of this information, we could
try to \emph{detect} when a shift has occurred. However, it is not
hard to see that, due to the noise in the rewards, detecting a shift
accurately would require $O(\sqrt{T})$ time steps.

\begin{figure}[t]
\centering
\includegraphics[scale=0.2]{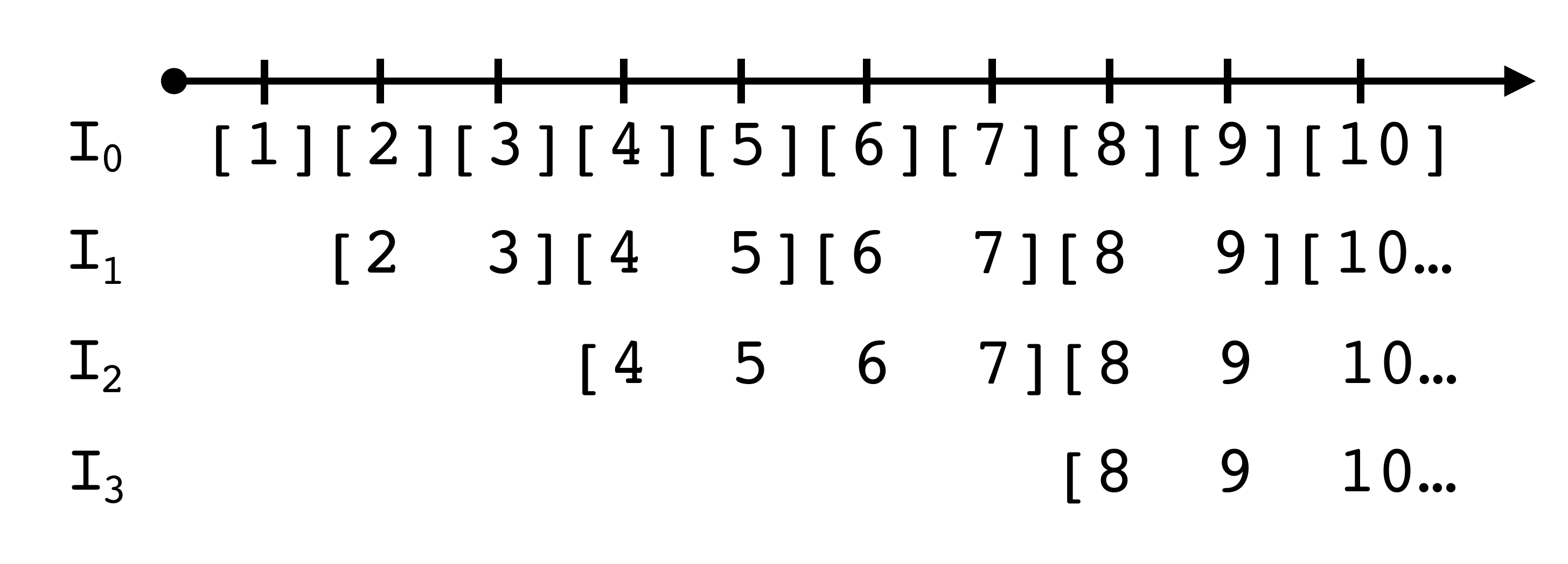}
\caption{Depiction of interval sets $\I_n$}
\label{fig:intervals}
\end{figure}

Instead, we propose the Follow-The-Best-Interval (FTBI)
algorithm. At a high level, FTBI maintains a small number of FTL instances, each
running over a different interval of a collection of nested
subintervals of $[T]$. The algorithm then follows the action
of the best performing FTL instance. 
More precisely, let
\begin{equation*}
\I_n =  \{[i\cdot 2^n, (i+1)\cdot 2^n-1]~:~i\in\N\}.
\end{equation*}
Here, $\I_n$ is a set of disjoint intervals of length $2^n$. Let $\I =
\bigcup_{n=0}^\infty I_n$ be the set of all such intervals. For every time $t$, define
$$
\ACT(t) := \{I\in\I~:~t\in I\}
$$
as the set of intervals in $\I$ that
contain $t$. It is immediate that $|\ACT(t)|  = O(\log t)$. A depiction
of the set $\I$ is shown in Figure~\ref{fig:intervals}. 

For every interval $I = [p,q] \in \I$, we keep  an instance of FTL
$F_I$. Let $$W_t(k, I) = \sum_{s=p}^t r_s(k) - r_s(x_s) $$ denote the
weight $F_I$ assigns to expert $k$ at time time $t$, where $x_t$ is
the expert, or version, chosen by our algorithm. Notice that this
differs slightly from  the traditional FTL definition as we subtract
the reward of the action taken by our algorithm. In doing so, we
ensure that the scale of the weights $W_t(k, I)$ is roughly the same for all
intervals $I$, regardless of their length.  

The expert and interval chosen by FTBI is given by
$$
(x_t, I_t) = \argmax_{k \in \{1,2\}, I \in \ACT(t)} W_{t-1}(k,I).
$$

A full description of FTBI can be found can be found in
Algorithm~\ref{alg:ftbi}.  The intuition for our algorithm is the
following: consider a segment $\T_i$ where rewards are
i.i.d.; due to the properties of the FTL algorithm, any FTL instance
beginning inside this segment will, in constant time, choose the
optimal action $k_i^*$. Therefore, as long as our algorithm chooses
FTL instances initialized inside a segment $\T_i$, FTBI will choose the
optimal action. We will show that the choice of weights $W_t(k, I)$
will guarantee that only instances of FTL that observe i.i.d rewards
will have large weights.

\begin{algorithm}
\caption{Follow-The-Best-Interval}\label{alg:mixedcoloring}
\begin{algorithmic}\label{alg:ftbi}
\STATE Initialize $W_0(k, I)=0$, for all $I\in\mathcal{I}$, $k\in\{1,2\}$.
\FOR {$t=1, 2,\ldots,T$}
\STATE Choose $(x_t, I_t)=\argmax_{\substack{k\in\{1,2\}, I \in\ACT(t)}} W_{t-1}(k, I)$
\STATE Observe reward vector $\r_t$
\STATE Update $W_t(k,I) = W_{t-1}(k, I)+r_t(k)-r_t(x_t)$, for all $I\in\ACT(t)$, $k\in\{1,2\}$.
\ENDFOR
\end{algorithmic}
\end{algorithm}

\paragraph*{Remark.} Since our algorithm is motivated by the A/B test problem, we focus on the case with two experts. However, the FTBI algorithm naturally extends to the setting with $K>2$ experts. In that case,
each interval needs to maintain $K$ weights, $W_t(k, I)$, $k=1,2,\ldots, K$. The algorithm chooses the expert and interval using:
$$
(x_t, I_t) = \argmax_{k \in [K], I \in \ACT(t)} W_{t-1}(k,I).
$$

\section{Analysis}
\label{sec:theory}

In this section we provide regret guarantees for our algorithm. The
main result of this paper is the following

\begin{theorem}
\label{thm:ftbi}
There exists a universal constant $C$ such that the shifting
pseudo-regret of FTBI is bounded by
$$
R_T \le \frac{C}{\Delta^3}N^2\log^3T.
$$
\end{theorem}

\subsection{Proof Sketch}
We begin by defining a disjoint cover of  each segment $\T_i$ by
elements  of $\I$. This cover was first introduced by 
\cite{daniely2015strongly}  to analyze their strongly adaptive
algorithm. 

\begin{theorem}[\cite{daniely2015strongly}, Geometric Covering]\label{thm:partition}
Any interval $I\subseteq [1,T]$ can be partitioned into two finite sequences of disjoint and consecutive intervals,
denoted by $I^{-m},\ldots,I^0\in \I$ and $I^1,\ldots,I^h\in\I$ such that
\begin{align*}
&(\forall~i\ge 1),~|I^{-i}|/|I^{-i+1}| \le 1/2,\\
&(\forall~i\ge 2),~|I^{i}|/|I^{i-1}| \le 1/2.
\end{align*}
\end{theorem}

\begin{figure}[t]
\centering
\includegraphics[scale=.35]{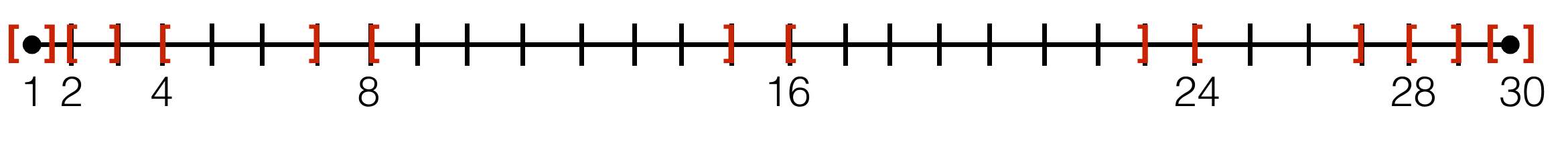}
\caption{Depiction of the cover.
We  show the covering of the interval $[1,30]$ by intervals $[1,1], [2,3], [4, 7], [8,15], [16,23], [24,27], [28,29], [30, 30] \in \I$.}
\label{fig:geometriccover}
\end{figure}

The above theorem shows that any interval $I$ in $[1,T]$ can be
partitioned by at most $2\log_2|I|$ intervals from $\I$. Since the
first sequence $I^{-m},\ldots,I^0$ has geometrically increasing
lengths and the second sequence $I^1,\ldots,I^h$ has geometrically
decreasing lengths, we call this partitioning the \emph{geometric
  covering} of interval $I$. A depiction of this cover can be seen in
Figure~\ref{fig:geometriccover}.

Denote by $\mathcal{C}\subseteq\I$ the collection of the geometric
coverings for all segments $\T_1,\ldots,\T_N$.  We know that,
$\mathcal{C}$ forms a partition of $[1,T]$, and there are at most
$2N\log_2 T$ intervals in $\mathcal{C}$. Furthermore, since each
interval $I\in\mathcal{C}$ is completely contained in $\T_j$ for some
$j$, the rewards are i.i.d. for any interval $I\in\mathcal{C}$.  The
main idea behind our proof is to analyze the regret of each interval
$I\in\mathcal{C}$, and to sum them up to obtain the overall regret of FTBI.
At a first glance bounding the regret for $I \in \mathcal{C}$ is
straightforward since the observed rewards are i.i.d. However, the FTL
instance $F_I$ over interval $I$ is not isolated. In particular,
it could  \emph{compete} with other instances $F_J$ over intervals $J
\supset I$. Therefore we must show that no
FTL instance accumulates a large weight in which case $F_I$ can
\emph{catch up} to the weights of $F_J$ in a small number of
rounds. 

For any $I = [p,q]\in \mathcal{C}$, let $\Long_I =\{J \in \I : J
\supseteq I\} $.  That is, $\Long_I$ consists of all intervals in $\I$
that contain $I$.  One can verify that, from the
definition of $\I$, if an interval $J=[p', q']\in\I$ overlaps with
$I$, i.e., $I\cap J \neq \emptyset$, and 
it was active before $I$,  i.e., $p' < p$, then it  must be that
$I\subset J$, i.e., $J\in \Long_I$. In view of this, it is the weight of the
intervals in $\Long_I$ that we need to bound. Let
\begin{equation*}
H^k_I = \argmax_{J \in\Long_I} W_{p-1}(k, J)
\end{equation*}
denote the interval in $\Long_I$ that assigns the highest weight to
action $k$ at the beginning of interval $I$.  Let $\cF_{t}$ denote the sigma algebra 
generated by all history up to time $t$ and  denote by $R_I$ the pseudo-regret of
FTBI over interval $I$ conditioned on all the previous history, i.e.,
$$
R_I:= \max_{k\in\{1,2\}}\EXP{\sum_{t=p}^q r_t(k)-r_t(x_t) \mid \cF_{p-1}}.
$$

\begin{theorem}\label{THM:FTBI-ONE-INTERVAL}
For  any $I=[p,q]\in\mathcal{C}$, we have
\begin{align}
R_I  \le & \EXPS{\max\{W_q(1, H_I^1) , W_q(2, H_I^2), 0 \}\mid \cF_{p-1}}  \nonumber\\
\le& \max\{W_{p-1}(1, H_I^1), W_{p-1}(2, H_I^2), 0\} + \frac{28}{\Delta^3}\log T+O(1). \label{eq:condi_maxweight}
\end{align}
\end{theorem}

We defer the full proof to the Appendix. This theorem shows that the
conditional regret in each interval 
depends on the maximum weight accumulated over all FTL instances
before time $p$  plus $\frac{28}{\Delta^3}\log T$. More
importantly, inequality~\eqref{eq:condi_maxweight} shows that at the
end of interval $I$ the maximum weight over all active FTL instances and all
actions can only increase by an additive constant. 

We can now prove Theorem~\ref{thm:ftbi}. First,
by construction of the geometric covering, we have $m := |\mathcal{C}| \le
2N\log_2T$. Let  $I(1), \ldots , I(m)$ be an
ordered enumeration of the intervals in $\mathcal{C}$. For every interval $I = [p,q]
\in \mathcal{C}$, let $W^*(I) = \max\{W_{p-1}(1, H_I^1), W_{p-1}(2,
H_I^2), 0\}$. Applying inequality~\eqref{eq:condi_maxweight}
recursively it is easy to see that:
$$
\EXPS{W^*(I(n))} \le (n-1)\left[\frac{28}{\Delta^3}\log T+O(1)\right],
$$
which again in view of \eqref{eq:condi_maxweight} yields
$$
\EXPS{R_{I(n)} }\le n\left[\frac{28}{\Delta^3}\log T+O(1)\right].
$$
Thus
$$
R_T = \sum_{n=1}^{m} \EXPS{R_{I(n)} } 
\le \frac{1}{2}m(m+1) \left[\frac{28}{\Delta^3}\log T+O(1)\right].
$$
The proof is completed using the fact that $m \le 2N\log_2T$.

\section{Experiments}\label{sec:experiments}

We conduct extensive simulations comparing our algorithm (FTBI) to AdaNormalHedge
(ANH) on both synthetic and real world datasets. In order to obtain an efficient implementation of ANH, we use the streaming trick mentioned in \cite{hazan2007adaptive,luo2015achieving}. 

\subsection{Synthetic Data}
The purpose of these simulations is to understand the
advantages of FTBI over ANH in a controlled environment for different parameters representing the
reward gap $\Delta$, number of shifts $N$, and round length $|\T_i|$. For each experiment we calculate the regret:
$ \overline{R}_T =  \sum_{i=1}^N \sum_{t=\tau_{i-1}+1}^{\tau_i} r_t(k_i^*) - r_t(x_t)$.
Each experiment is replicated 20 times and we report the average result.
The error bars represent one
standard deviation. For every experiment the rewards of the two experts over
the $i$-th segment  $\T_i$ are sampled i.i.d. from a Bernoulli distribution with parameters $p_{i,1}, p_{i,2}$.

\begin{figure}[h]
\centering
\begin{tabular}{cccc}
\includegraphics[scale=0.6]{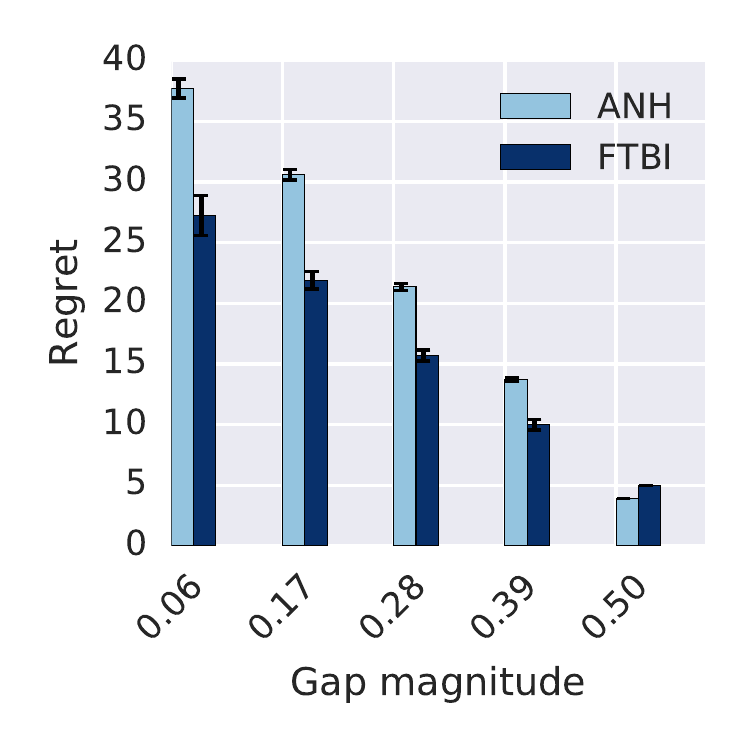}  &
\includegraphics[scale=0.6]{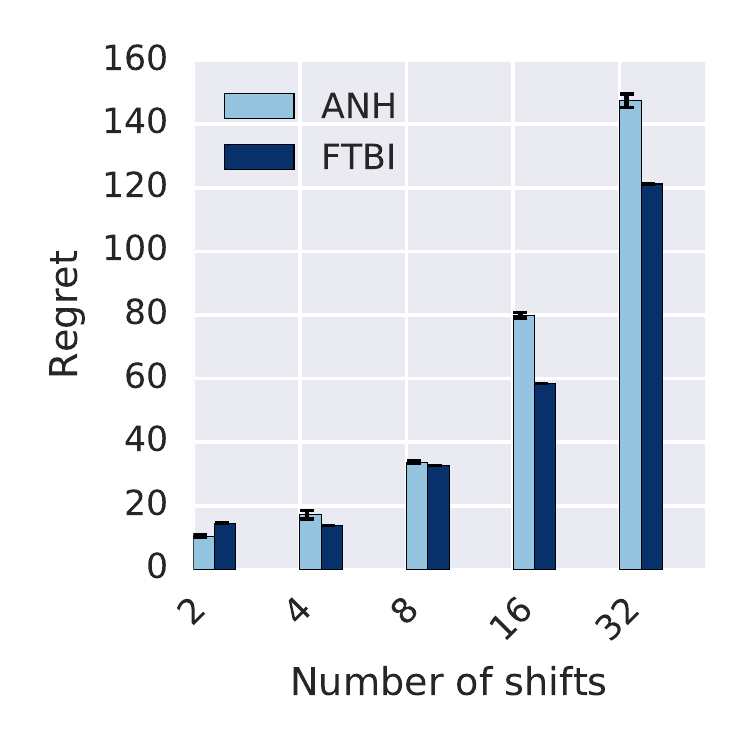} \\
(a) & (b) \\
\includegraphics[scale=0.6]{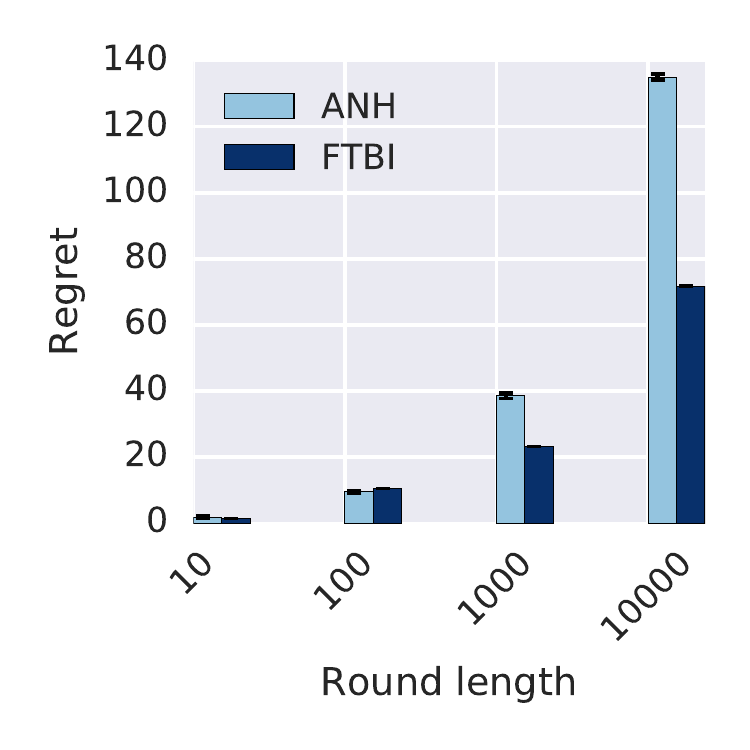} &
\includegraphics[scale=0.58]{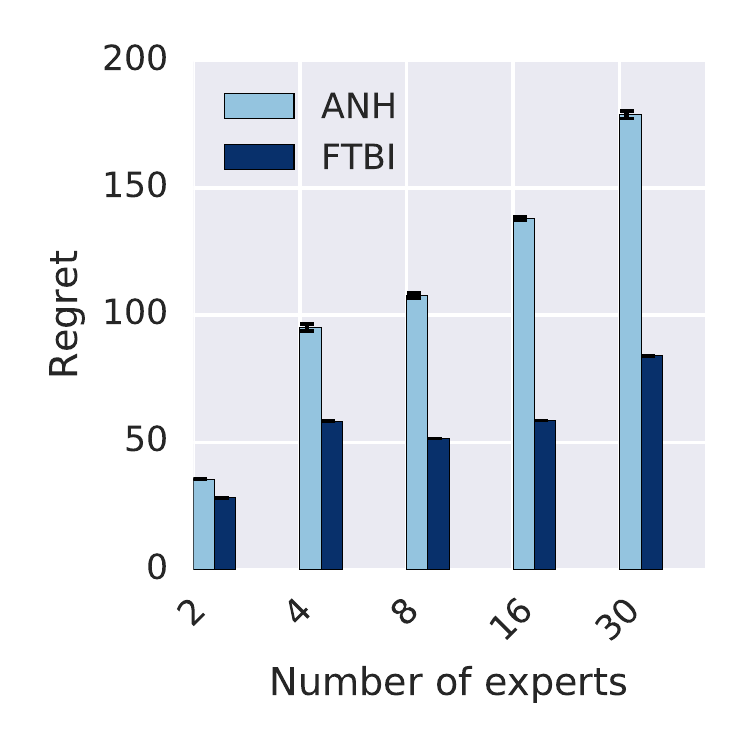} \\
 (c) & (d)
\end{tabular}
\caption{Performance on synthetic data. Regret as a function of (a) reward gap $\Delta$, (b) number of
shifts $N$, (c) round length $|\T_i|$. (d) number of experts $K$.}
\label{fig:results}
\end{figure}

\begin{figure}[h]
\centering
\begin{tabular}{ccc}
\includegraphics[scale=0.6]{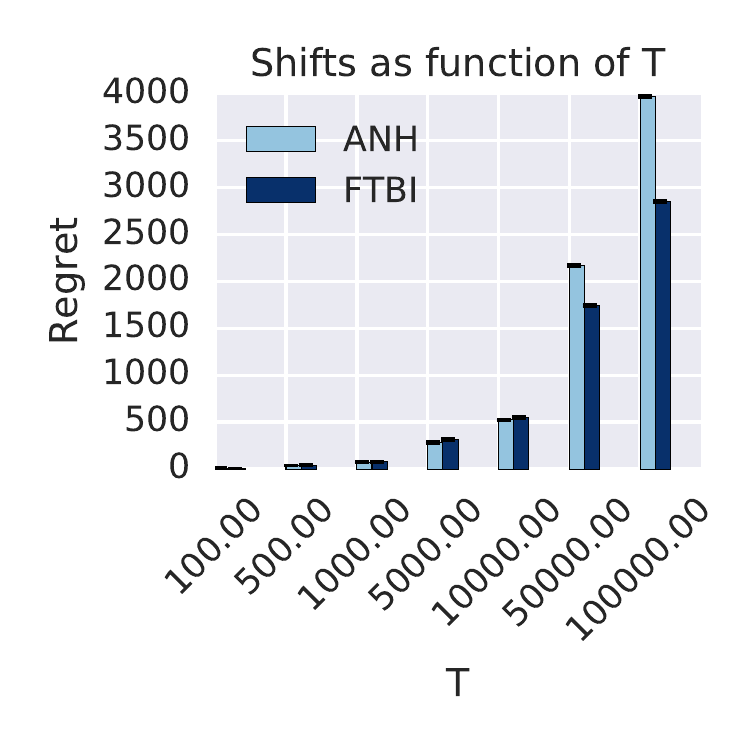} &
\includegraphics[scale=0.6]{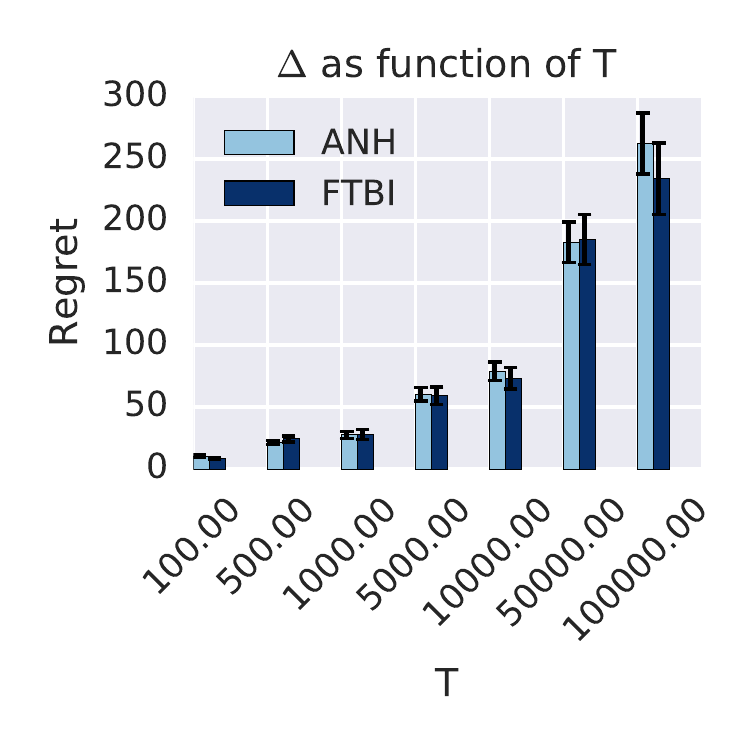} &
\includegraphics[scale=0.6]{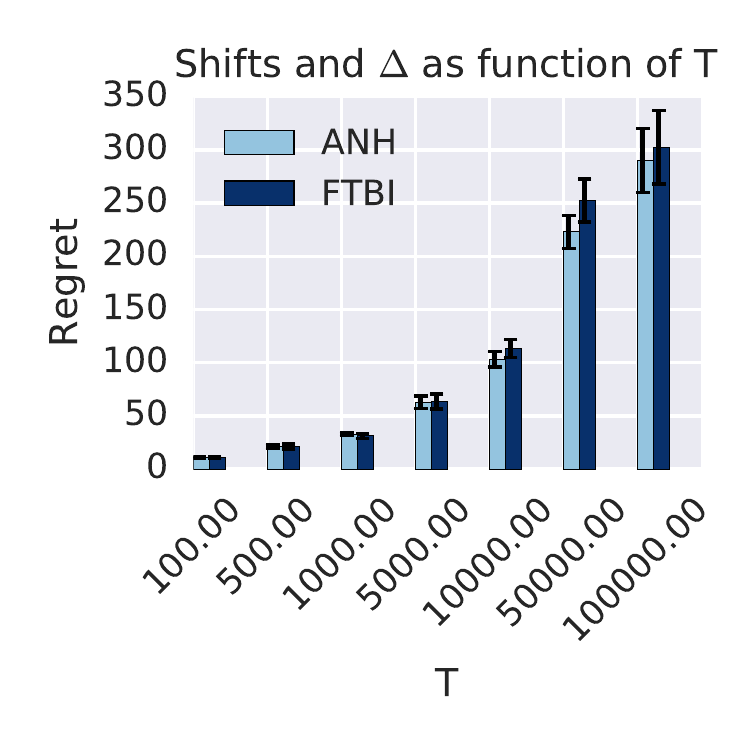} \\
(a) & (b) & (c) 
\end{tabular}
\caption{Performance on synthetic data. Regret as a function of number of shifts $N$ and reward gap $\Delta$, with $N$ and $\Delta$ chosen as functions of $T$. (a) Fix $\Delta = 0.1$ and let $ N = \sqrt{T}$. (b) Fix $N=10$ and let $\Delta = 1/\sqrt{T}$. (c) Let $ N = \sqrt{T}, \Delta = 1/\sqrt{T}$.}
\label{fig:results_function_of_T}
\end{figure}

We explore the performance of FTBI as a function of the reward gap, $\Delta$, the number of shifts, $N$, round length, $T_i$, and the number of experts, $K$. 

\subsubsection{Sensitivity to the reward gap $\Delta$} For this experiment, we vary the reward gap over the set $\Delta \in \{
  0.06, 0.17, 0.28, 0.39, 0.5\}$ and  choose parameters $p_{i,k}$ such that
  $|p_{i,1} - p_{i,2}| = \Delta$. We allow for two intervals
  $\T_i$ with $800$ rounds per interval, and plot the results in Figure ~\ref{fig:results}(a). Observe that FTBI consistently outperforms ANH, and the regret exhibits a significantly better dependence on $\Delta$ than guaranteed by Theorem \ref{thm:ftbi}. 

\subsubsection{Sensitivity to the number of shifts, $N$} We vary the number of shifts $N$ from $2$ to $32$. For each setting, we sample parameters $p_{i,1}$, $p_{i,2}$ 
from a uniform distribution in $[0,1]$, and fix the number of rounds per interval to $800$. Observe that a new draw of $p_{i}'s$ initiates a shift with high probability as
$p_{i,k}\neq p_{i+1,k}$. We compare the regret of FTBI and ANH in Figure~\ref{fig:results}(b).
Notice that although our theoretical bound
shows a quadratic dependence on the number of shifts, we do not see
this unfavorable bound empirically, and again FTBI consistently
outperforms ANH.

\subsubsection{Sensitivity to the round length, $T_i$}  We fix the number of intervals to $2$. On the first interval expert $1$ has mean reward $0.7$ and expert $2$ has reward $0.5$.  The means are swapped in the second interval. We vary the length of each interval over the set $\{10^1, 10^2, 10^3, 10^4\}$, and show the results in Figure~\ref{fig:results}(c). This experiment captures the main downside of ANH. Specifically, ANH is slow to adjust to a shift in  distributions. Indeed, the longer one arm dominates the other, the longer it takes ANH to switch when a distribution shift occurs. By contrast, FTBI adjusts much faster. 

\subsubsection{Sensitivity to  the number of experts, $K$}
As mentioned in Section~\ref{sec:algorithm}, although our main focus is the A/B test problem, as an expert learning algorithm, FTBI also applies to scenarios where there are $K>2$ experts. In this experiment, we study the performance of FTBI when we vary the number of experts $K$. We sample the  parameters $p_{i,k} (k=1,2,\ldots, K)$  i.i.d.  from a uniform distribution
in $[0,1]$. We fix the length of each segment $\T_i$  to $800$ rounds, and vary the number of experts $k$ from $2$ to $32$. We show the results in Figure~\ref{fig:results}(d). Although FTBI does not have theoretical guarantees in this setting, it consistently outperforms
ANH. In addition, FTBI appears to scale better than ANH as the number of experts grows. 

\subsubsection{Sensitivity to the number of shifts and reward gap as functions of $T$}
We also test the performance of FTBI and ANH in the settings where the number of shifts $N$ and/or the reward gap $\Delta$ are functions of $T$. These settings simulate practical scenarios where there are relatively large number of shifts and ) small reward gap between the optimal and sub-optimal experts. More importantly, it is in these scenarios that our theoretical bounds become trivial as setting $N= \sqrt{T}$ or $\Delta = \frac{1}{\sqrt{T}}$ makes the bound of Theorem~\ref{thm:ftbi} simply $O(T)$. In theory our algorithm should perform worse than ANH, However, as seen in Figure~\ref{fig:results_function_of_T} this is not the case. For Figure~\ref{fig:results_function_of_T}(a), we fixed $\Delta = 0.1$ and let $ N = \sqrt{T}$. We observe that FTBI achieves significantly smaller regret than ANH. Again, this is due to the fact that larger times without shifts are detrimental to the performance of ANH. For Figure~\ref{fig:results_function_of_T}(b), we fix $N=10$ and let $\Delta = 1/\sqrt{T}$ since this setup corresponds to experts with similar performance both algorithms achieve similar regret.


\begin{figure}[t]
\centering
\begin{tabular}{cc}
\includegraphics[scale=0.6]{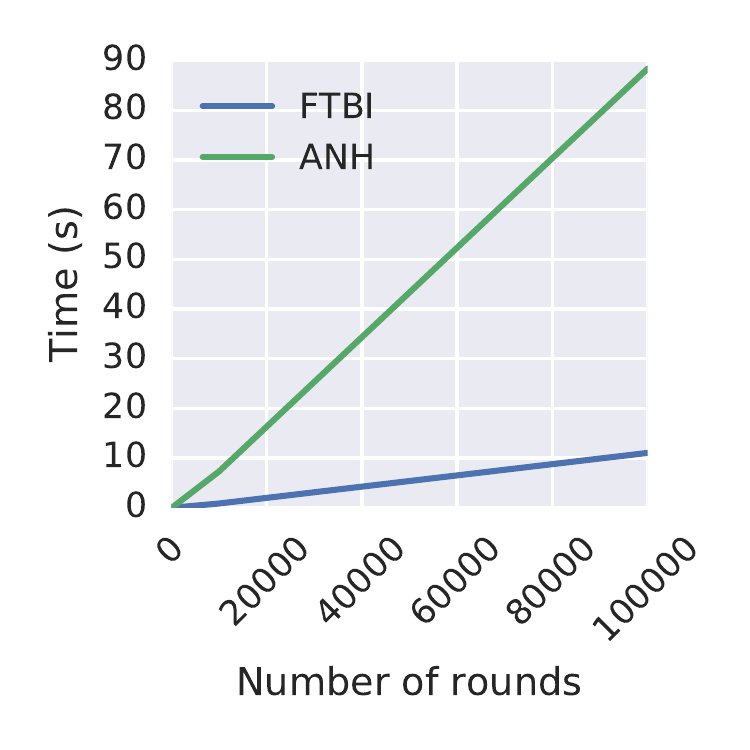} & \includegraphics[scale=0.6]{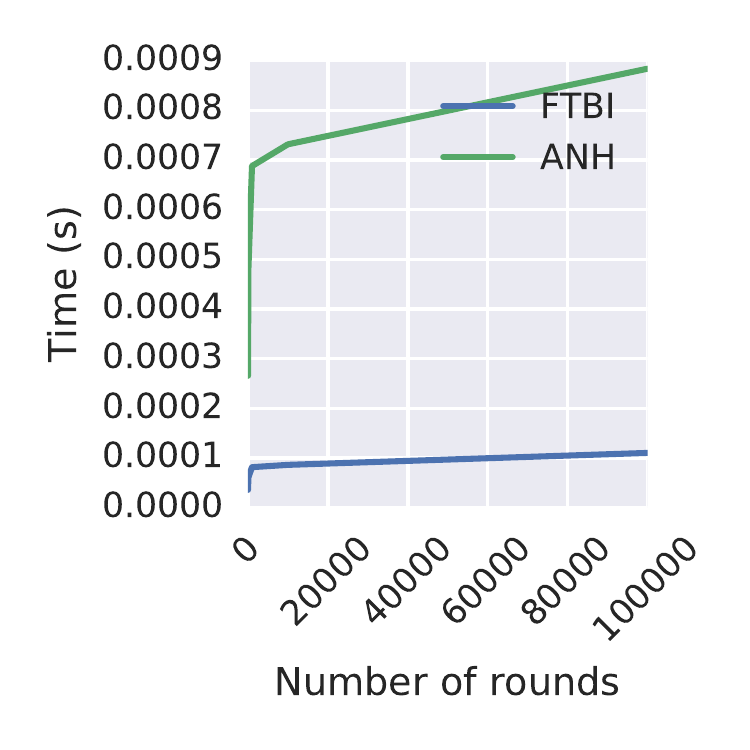} \\
(a)  & (b)  \\
\end{tabular}
\caption{Performance on synthetic data. Running time (in seconds) of FTBI and ANH. (a) Total running time. (b) Time per round.}
\label{fig:time_perf}
\end{figure}

\subsubsection{Scalability } To conclude this synthetic evaluation, we compare the running time
 of both algorithms. We vary the length of each round over the set $\{10 ,100 ,1000, 10000, 100000\}$ and show the results in Figure~\ref{fig:time_perf}. Notice that while both algorithms have total running time grow linearly, and the per round time increases logarithmically, FTBI is approximately 10 times faster across the board. 

To conclude, while ANH has a stronger theoretical regret bound, FTBI consistently performs better than ANH in the experiments. We believe that our current regret analysis can be further improved so that the theoretical regret bounds can match the performance of FTBI in practice. We propose this direction in our future work.

\subsection{Evaluation on Public Datasets}
\label{section:publicdata}
We now evaluate our algorithm by simulating the outcome of an A/B test based on public time series data. The methodology is the following: given a time series we consider two prediction systems. The first one predicts that the value will be above a fixed threshold $\tau$ and the other one predicts the it  will be below $\tau$. The reward of each system is $1$ when the system is correct and $0$ otherwise. The goal is to find a combination of systems that yields the largest cumulative reward. As mentioned in the introduction, this setup matches that of expert learning with two experts. 

The first time series we consider is the air quality data set (\url{https://archive.ics.uci.edu/ml/datasets/Beijing+PM2.5+Data}) consisting of hourly measurements of air quality in Beijing. The threshold of our systems is given by  $\tau = 75$ corresponding to the rounded median of  the air quality index. The second data set consists of measurements of electric power usage (\url{https://archive.ics.uci.edu/ml/datasets/Individual+household+electric+power+consumption}) and the threshold is given by $\tau=0.5$. 

We compare the performance of FTBI, ANH, and FTL algorithms in these tasks. Before  presenting the results of these comparisons we first show why these data sets benefit from the adaptive nature of FTBI. In Figure~\ref{fig:threshold_results}(a)(b), we plot the difference of the cumulative reward of the thresholding systems. Notice that if one system consistently dominated the other (a favorable case for FTL) we would see a monotonically increasing curve. Instead, we observe an oscillatory behavior, which indicates that the best system changes several times over the time period. 

We run the three algorithms in these data sets and show the total reward of FTBI and ANH as a relative improvement over FTL in Figure~\ref{fig:threshold_results}(c)(d). There, we see that both ANH and FTL largely outperform FTL with improvements of up to $85\%$, which validates the fact that these algorithms are able to adapt to the changes in the best system. Furthermore, our algorithm outperforms ANH. In the first data set it is considerably better, while in the second one the difference is not as large. By looking at the corresponding cumulative reward plots we see that the first task is noisier than the second one which leads us to believe that FTBI performs better under noisy conditions.

\begin{figure}[t]
\centering
\begin{tabular}{cccc}
\includegraphics[scale=0.6]{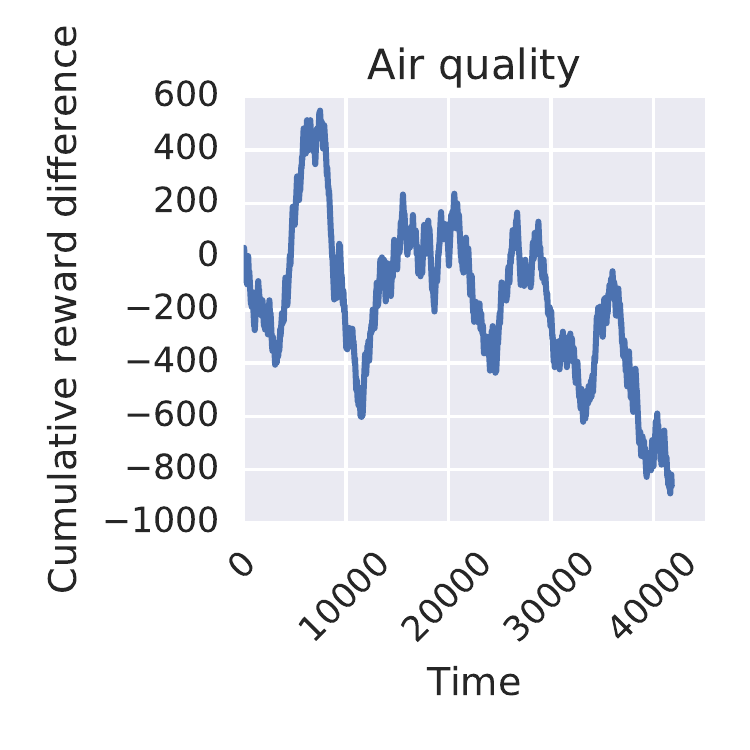}  &
\includegraphics[scale=0.6]{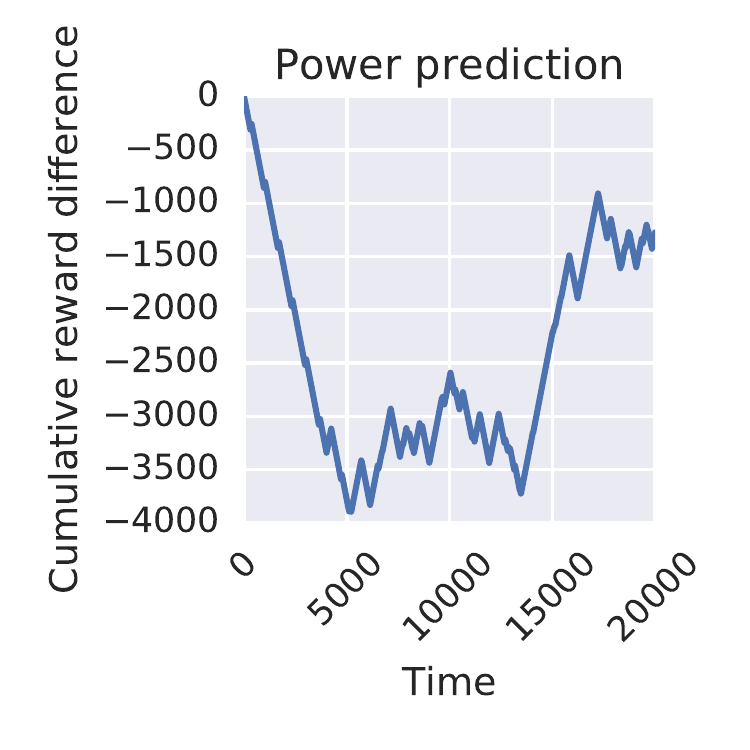} \\
(a) & (b) \\
\includegraphics[scale=0.6]{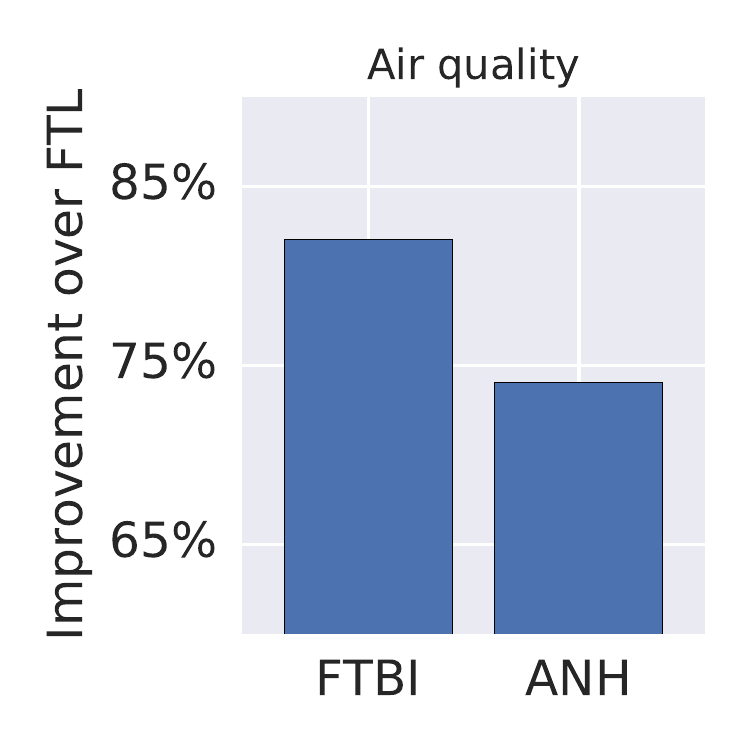} &
\includegraphics[scale=0.6]{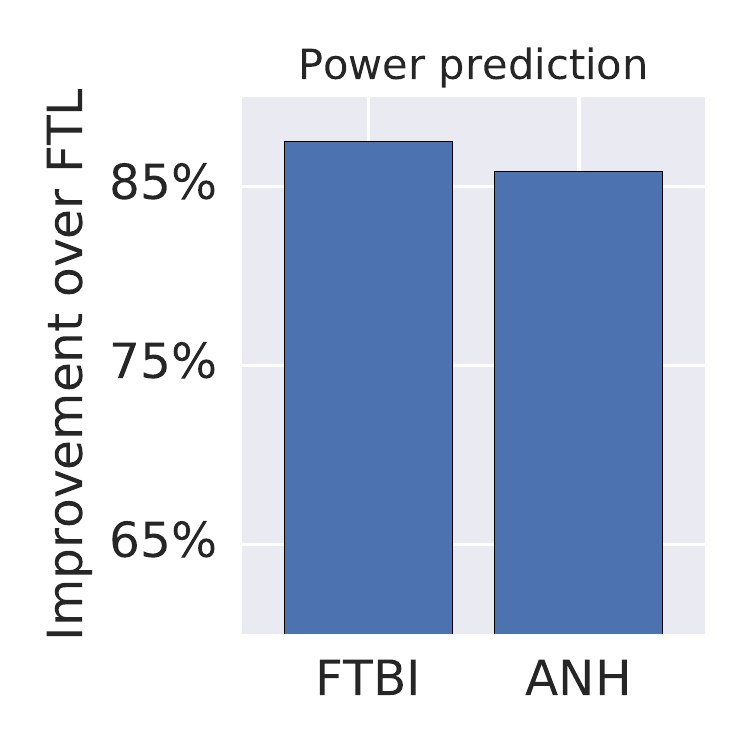} \\
 (c) & (d)
\end{tabular}
\caption{(a)(b) Difference of cumulative rewards of both systems. (c)(d) Total reward of FTBI and ANH as a relative improvement over the total reward of FTL.}
\label{fig:threshold_results}
\end{figure}

\ignore{
\begin{figure}[t]
\begin{tabular}{cc}
(a) \includegraphics[scale=0.35]{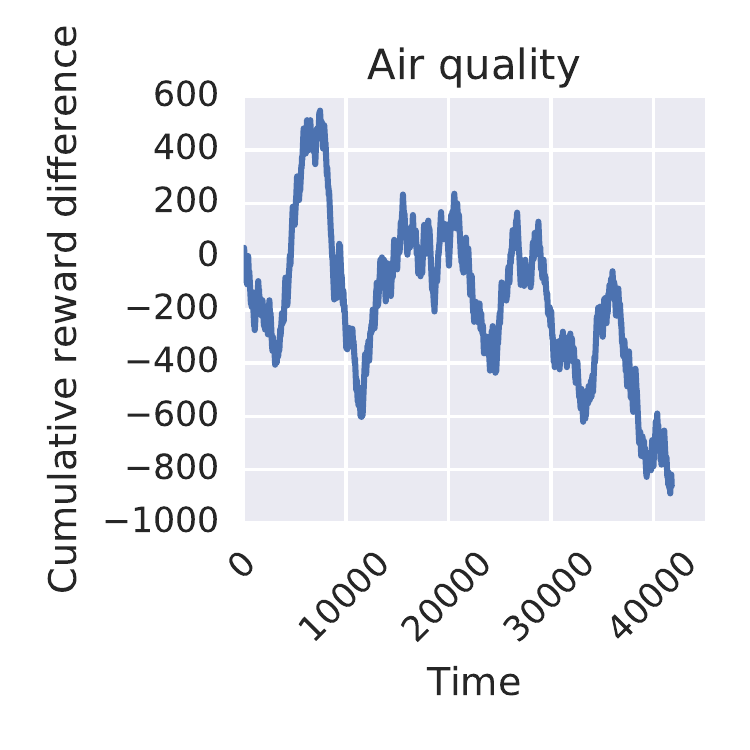}
&
\includegraphics[scale=0.35]{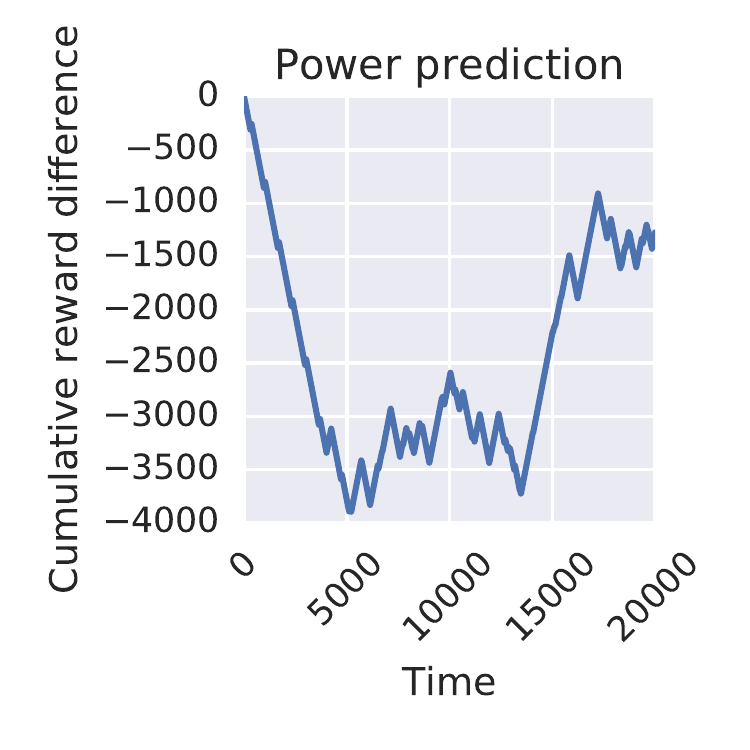} \\


(b) \includegraphics[scale=0.35]{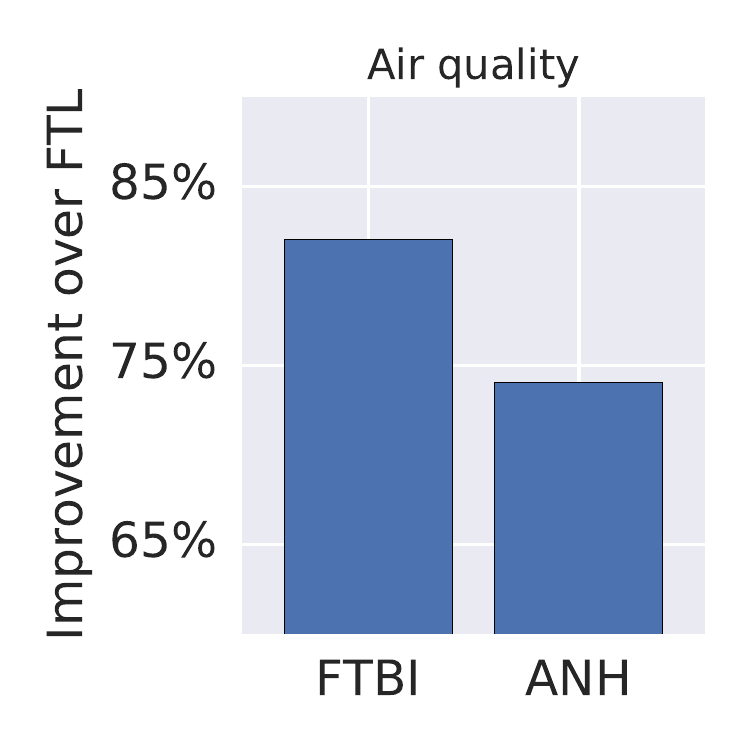}
&
\includegraphics[scale=0.35]{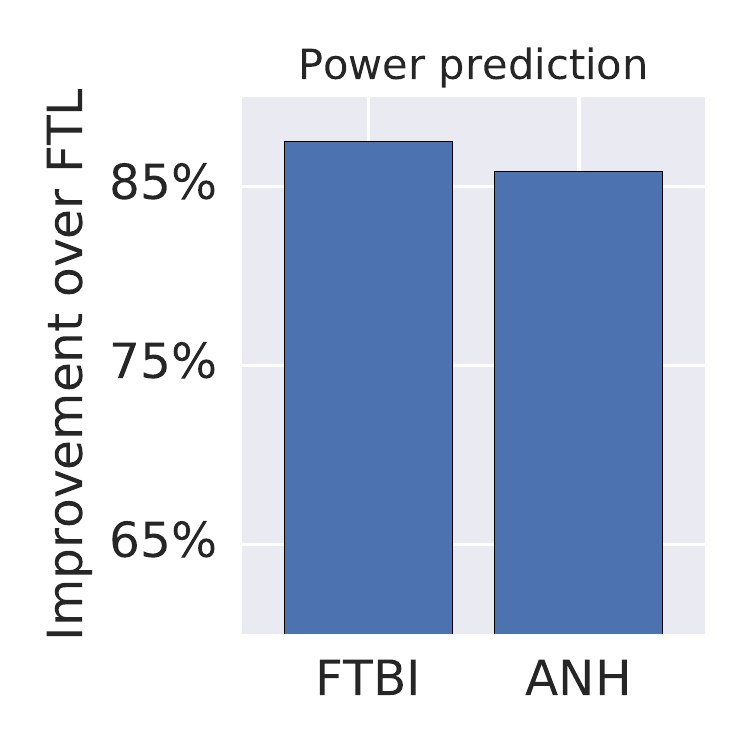}
\end{tabular}
\caption{(a) Difference of cumulative rewards of both systems. (b) Total reward of FTBI and ANH as a relative improvement over the total reward of FTL.}
\label{fig:threshold_results}
\end{figure}
}
\subsection{Evaluation on an Advertising Exchange Platform}

Finally, to showcase the advantage of our method in practice, we apply the FTBI algorithm to compare two pricing mechanisms on an advertising exchange platform using real auction data. We consider the
problem of setting reserve (or minimum) prices in the auctions, with the goal of increasing revenue. Learning good reserve prices is a notoriously challenging problem, with a myriad of algorithms designed to solve this task ~\cite{gentilebianci,gentilemansourbianci,munozvassilvitskii,mohrimunoz}. 

In this experiment, we collect a sample of auctions (highest and second highest bid) for
an eight day window from four different publishers, see the total
number of auctions in Table~\ref{table:stats}. We consider an A/B test over two algorithms
$\cA_1$ and $\cA_2$ for setting the reserve price; we treat them as
black boxes for the purpose of the experiment.
At every time $t$, we
can compute the revenues\footnote{The revenue in a particular auction equals the amount of money that the winner of the auction pays to the publisher.} $r_t(1)$ and $r_t(2)$ that could be obtained by each of these algorithms. (We ignore the question of incentive compatibility of learning in auctions, as it is beyond the scope of this work.) 

\begin{table}[h]
\centering
\begin{tabular}{|c|c|c|c|}
\hline
\small{Publisher} & \small{Number of auctions} 
&  \small{Publisher} & \small{Number of auctions} \\ 
\hline
\small{A} & \small{20,745}  
&
\small{B} & \small{115,341} \\
\hline
\small{C} & \small{64,524} 
&
\small{D} & \small{34,789} \\
\hline
\end{tabular}
\caption{Total number of auctions in our data set}
\label{table:stats}
\end{table}

\begin{figure}[t]
\centering
\begin{tabular}{cc}
\includegraphics[scale=0.5]{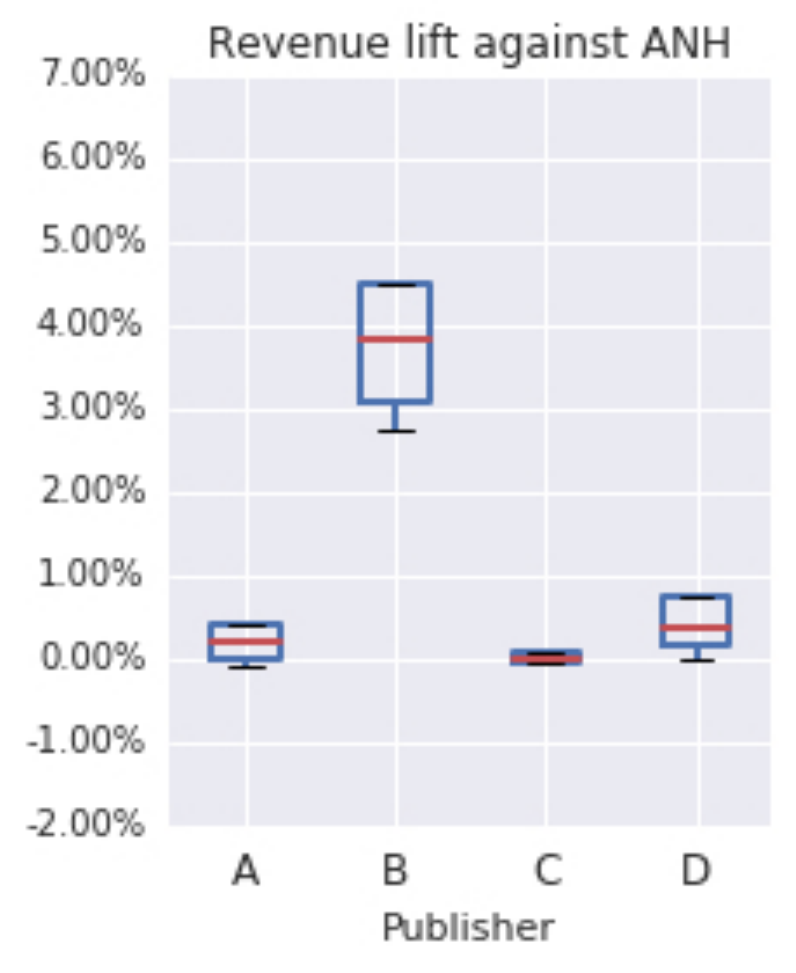}
&
\includegraphics[scale=0.5]{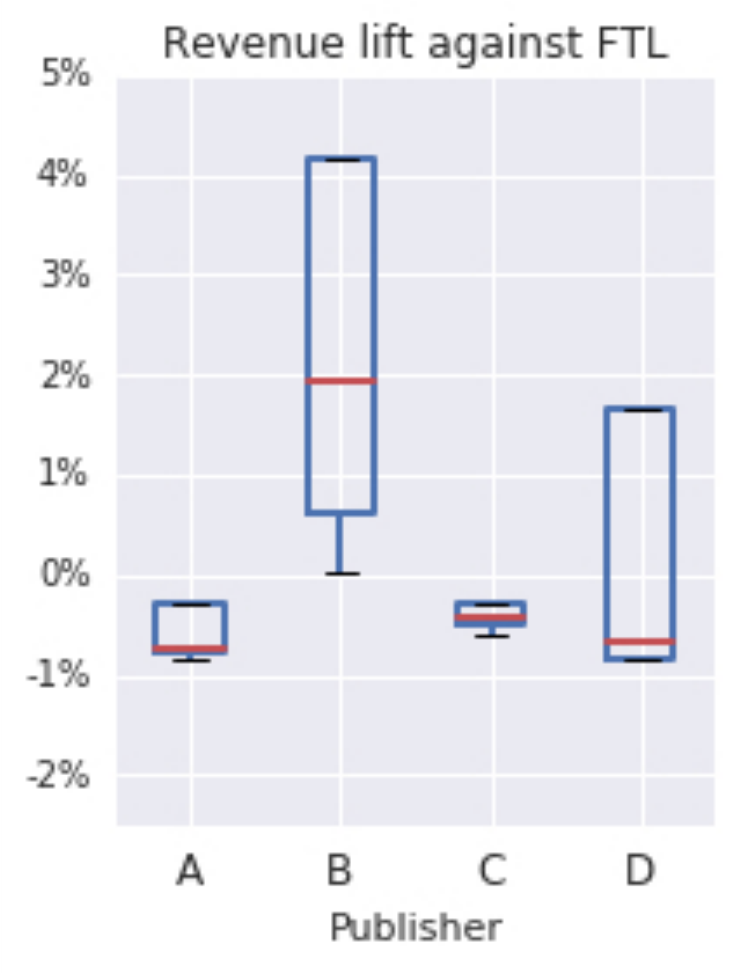} \\
(a) & (b)
\end{tabular}
\caption{Performance on Auction data. Revenue lift comparisons in percent improvement. (a) Lift of FTBI against ANH. (b) Lift of FTBI against FTL. The box limits represent the 25 and 75 percent quantiles while the
whiskers represent the 10 and 90 percent quantiles.}
\label{fig:lift}
\end{figure}

\begin{figure}
\centering
\includegraphics[scale=0.45]{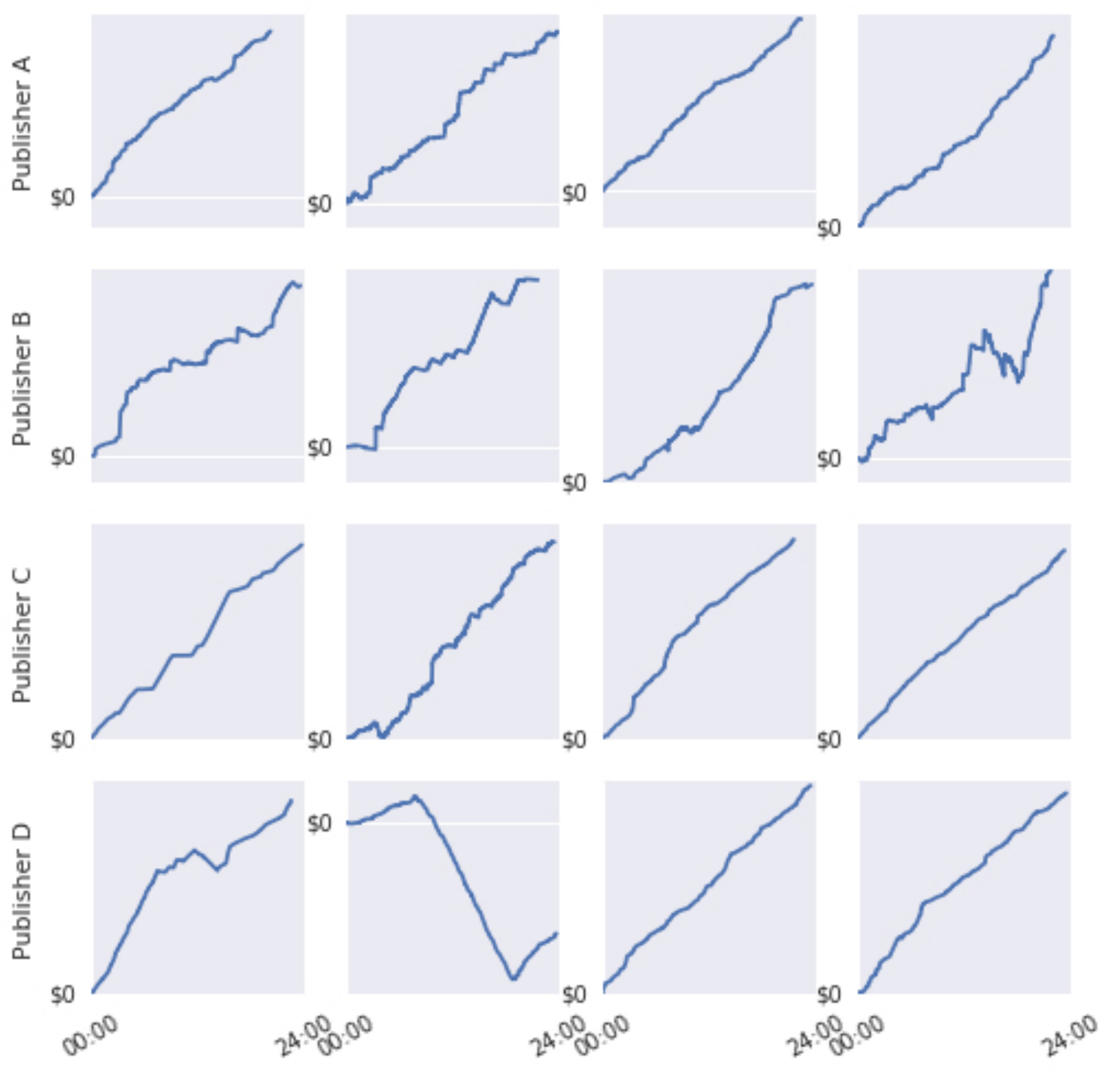}
\caption{Performance on Auction data. Difference in the cumulative revenue of algorithms $\cA_1$ and $\cA_2$.}
\label{fig:rev-diff} 
\end{figure}

Since the reserve price must be set prior to running the
auction, our goal is to select one of these algorithms at each point
of time $t$, to maximize the overall revenue. We run the Follow-The-
Leader (FTL) algorithm, AdaNormalHedge (ANH) and Follow-The-Best-Interval (FTBI)
over these sequences of rewards. We present the
relative revenue lift of FTBI (defined as the relative difference in revenue) 
over the two other algorithms in Figure~\ref{fig:lift}, where we treat
 each day as a separate experiment. The
box limits represent the 25 and 75 percent quantiles while the
whiskers represent the 10 and 90 percent quantiles. Observe from Figure~\ref{fig:lift}(a) is that FTBI is
consistently better than ANH, in some cases improving total revenue by
more than 4\%. 

Figure~\ref{fig:lift}(b) shows that FTBI does not always compare
favorably with respect to FTL. This can explained by the
fact that in these sequences, one algorithm is consistently better
than the other throughout the day. It is well known that FTL has the
best regret guarantees in this, non-switching, scenario, and FTBI essentially matches its
the performance on publishers $A$ and $C$. In cases where the optimum strategy
switches between $\cA_1$ and $\cA_2$, FTBI tends to outperform
FTL, sometimes significantly.

To investigate this further, in Figure~\ref{fig:rev-diff}, we plot the
difference in the cumulative revenue of $\cA_1$ and $\cA_2$ as a
function of time for each publisher for four different days.  When the
time series increases, it represents a 
segment of time where $\cA_1$ was consistently better than $\cA_2$;
the opposite is true when the series decreases. In particular, notice 
that, for publishers $A$ and $C$, $\cA_1$ consistently outperforms
$\cA_2$, explaining the results in Figure~\ref{fig:lift}(b). On the
other hand, the best algorithm in publisher $B$ switches considerably
in days 2 and 4 and there is a clear switch for publisher $D$ in day
2. It is in these cases that we obtain the most benefits from an
adaptive algorithm such as FTBI. 

\section{Conclusion}
In this work, we study the setting of monitoring non-stationary A/B tests, and formulate an expert learning framework for this problem. We develop the Follow-The-Best-Interval (FTBI) algorithm and prove theoretical bounds on its regret. Our empirical evaluation suggests that it achieves lower regret than the state-of-the-art algorithm across many synthetic and real world datasets. In addition, FTBI is  much faster than the state-of-the-art AdaNormalHedge algorithm. We suggest improving the theoretical guarantee of FTBI as a future research direction.

%% file: proofs.tex
\section*{Appendix}
\appendix

\section{Proof of Theorem~\ref{THM:FTBI-ONE-INTERVAL}}

To simplify notation we fix $I=[p,q] \in
\mathcal{C}$ and let $H^k:= H_I^k$ for $k = 1,2$. We define
$W_1:=W_{p-1}(1, H^1)$ and $W_2:=W_{p-1}(2, H^2)$, we also let $\E$
denote the conditional expectation with respect to
$\cF_{p-1}$, and $\mathbb{P}$ denote the conditional probability with respect to
$\cF_{p-1}$. Finally, we assume, without loss of generality
that expert 1 is optimal over the segment $I$. 

The proof of Theorem~\ref{THM:FTBI-ONE-INTERVAL} depends on the
following three technical lemmas.
\begin{lemma}
\label{LEM:CASE1}
If $W_1 \ge W_2$, we have
\begin{equation}
\label{eq:w1gtw2}
\begin{aligned}
R_I \le  \EXPS{\max\{W_q(1, H^1), W_q(2, H^2), 0 \} - W_1 } \le \frac{6}{\Delta^2} \log T + O(1).
\end{aligned}
\end{equation}
\end{lemma}

\begin{lemma}\label{LEM:CASE2}
If $W_2 > W_1$, we have
\begin{equation}
\label{eq:w1ltw2}
\begin{aligned}
R_I \le \EXPS{\max\{W_q(1, H^1), W_q(2, H^2), 0 \} - W_1 } \le \Big(1+\frac{2}{\Delta}\Big)(W_2 -W_1)
+\frac{16}{\Delta^2} \log T + O(1).
\end{aligned}
\end{equation}
\end{lemma}

\begin{lemma}\label{LEM:CASE3}
If $W_2 - W_1 > \frac{6}{\Delta^2}\log T$, we have
\begin{equation}
\label{eq:w1ltw2_v2}
\begin{aligned}
R_I \le \EXPS{\max\{W_q(1, H^1) , W_q(2, H^2), 0 \} - W_1} \le W_2 - W_1 +\frac{6}{\Delta^2}\log T +O(1).
\end{aligned}
\end{equation}
\end{lemma}

The proof of Theorem~\ref{THM:FTBI-ONE-INTERVAL} is now immediate.
Combining Lemma~\ref{LEM:CASE2} and Lemma~\ref{LEM:CASE3}, we know
that for $W_2 > W_1$, we always have
\begin{align}
R_I &\le \EXPS{\max\{W_q(1, H^1), W_q(2, H_2), 0 \} - W_1}  \nonumber\\
&\le W_2 - W_1 + (\frac{12}{\Delta^3} + \frac{16}{\Delta^2}) \log T
  +O(1)  \nonumber \\
&\le W_2 - W_1 + \frac{28}{\Delta^3}\log T+O(1). 
\end{align}
Moreover, by definition of $H^1$ and since $I \in \Long_I$ we must
have $W_1 \geq W_{p-1}(1, I) = 0$. Therefore by 
Lemma~\ref{LEM:CASE1}, Lemma~\ref{LEM:CASE2}  and the fact that
$W_1$ is $\cF_{p-1}$ measurable we have:
\begin{align*}
  R_I & \leq R_I + W_1 
 \leq \EXPS{\max \{W_q(1, H^1), W_q(2, H^2), 0 \} } \\
& \leq \max\{W_1, W_2, 0\}  +  \frac{28}{\Delta^3}\log T+O(1),
\end{align*}
where the last inequality follows from \eqref{eq:w1gtw2} and \eqref{eq:w1ltw2}.

\section{Proof of Lemma~\ref{LEM:CASE1}}
By definition of regret we have
\begin{align*}
R_I &= \EXP{\sum_{t=p}^q r_t(1) - r_t(x_t)} = \EXPS{W_q(1, H^1) - W_1} \\
&\le \EXPS{\max\{W_q(1, H^1), W_q(2, H^2), 0 \} - W_1 }:=\tilde{R}_I.
\end{align*}

Fix $\delta >
0$, and let $\tau = \inf\{q >  t \geq p : W_t(1, H^1) \geq W_1 + \delta\}$,
and $\tau = \infty$ if $W_t(1, H^1) < W_1 + \delta$ for all $t < q$.
Define the following two events: 
$$ 
X:=\{W_q(1, H^1) \ge W_q(2, H^2)\},
$$
$$
Y:=\{\tau < q \ \wedge \ x_t = 1 \ \forall\, t \in [\tau + 1, q]  \}
\cup \{\tau = \infty\}.
$$
A simple observation is that, if $Y$ happens, we must have $W_t(1, H^1)
\le W_1+\delta + 1$ for all $t\in I$. Indeed, for $t < \tau$, $W_t(1, H^1)
< W_1+\delta$ by definition. Thus $W_\tau(1, H^1) \le W_1 + \delta+1$, and then
$W_t(1, H^1)$ remains unchanged as $x_t = 1$ for all $t \in [\tau + 1,q]$.

Therefore, conditioned on the event $X\cap Y$, we always have
$$
\max\{W_q(1, H^1), W_q(2, H^2), 0 \} - W_1 \le \delta+1.
$$
Therefore we can bound $\tilde{R}_I$ as 

\begin{align*}
\tilde{R}_I = & \EXPS{\max\{W_q(1, H^1), W_q(2, H^2), 0 \} - W_1 \mid X\cap Y}\probs{X\cap Y}\\
& + \EXPS{\max\{W_q(1, H^1), W_q(2, H^2), 0 \} - W_1 \mid \bar{X}\cup\bar{Y}}\probs{\bar{X}\cup\bar{Y}} \\
\le& \delta + 1 + 2T\probs{\bar{X}\cup\bar{Y}}.
\end{align*}

Moreover, since $W_1 \ge W_2$, we have
$$
\probs{\bar{X}} 
\le \prob{\sum_{t=p}^qr_t(1)-r_t(2) < 0} \le e^{-\frac{1}{2}\Delta^2\abss{I}},
$$
where we have used the fact that $ \EXPS{r_t(1) - r_t(2)} \geq \Delta$ and
Hoeffding's inequality. 

Let us bound $\probs{\bar{Y}}$. Notice that event $Y$ does not happen
if and only if $\tau < q$ and there exists $\tau' > \tau$ such that
$x_{\tau'} = 2$ and $x_t = 1$ for $t \in [\tau+1, \tau'-1]$. Then $W_t(1,
H^1)$ remains unchanged for $t=\tau, \ldots, \tau'-1 $ and $W_t(1, H^1) \ge W_1 + \delta$ for every $t  \in [\tau, \tau'-1]$. Thus, the previous
event can happen if and only if there exists $t \in [p,q]$ such that
$W_{t-1}(1, H^1) \ge W_1 + \delta$ and $x_t  = 2$.  Therefore, let 
$$
E_t := \{W_{t-1}(1, H^1) \ge W_1+\delta~\wedge~ x_t=2\},
$$
for all $t\in[p+1,q]$. It is easy to see that $\bar{Y} =
\cup_{t=p+1}^q E_t$. Also, if  $t < p+\delta$,  then $W_{t-1}(H^1)
< W_1+\delta$. Therefore, we need only bound the probability of $E_t$ for $t \ge
p+\delta$. Recall in Algorithm~\ref{alg:ftbi}, we define
\begin{equation}\label{eq:def_it}
(x_t, I_t)=\argmax_{\substack{k\in\{1,2\}, \tilde{I} \in\ACT(t)}} W_{t-1}(k, \tilde{I}).
\end{equation}
Therefore, we have
\begin{align*}
 \probs{E_t} = & \probs{W_{t-1}(1, H^1) \ge W_1+\delta, x_t = 2} \\
 \leq & \probs{I_t = H^2, x_t =2} + \sum_{\substack{J\in\ACT(t) \\ p_J > p}}\probs{W_{t-1}(1,
  H^1) \ge W_1+\delta ,
 I_t =  J, x_t=2} \\
\le & \probs{W_{t-1}(1, H^1)  < W_{t-1} (2, H^2)} \\
 &+ \sum_{\substack{J\in\ACT(t) \\ p_J>p}}\probs{W_{t-1}(1, J) <
  W_{t-1}(2, J), W_{t-1}(2, J) \ge \delta},
\end{align*}

where we denote by $p_J$ the starting time of interval $J$.
Applying Hoeffding's inequality, we can bound the first term by 
$e^{-\frac{(t - p)\Delta^2}{2}} \leq e^{-\frac{\delta
    \Delta^2}{2}}$. For the second term notice that if $t - p_J <
\delta$ then the probability of the summand is $0$. Thus we can
restrict ourselves to intervals for which $t - p_J \ge \delta$  and we
can bound the second term by
\begin{align*} 
 \sum_{\substack{J\in\ACT(t) \\ p_J>p , t - p_J \ge
  \delta}}\probs{W_{t-1}(1, J) < W_{t-1}(2, J)}  
\le   \sum_{\substack{J\in\ACT(t) \\ p_J>p, t-p_J\ge \delta}} 
e^{-\frac{(t-p_J)\Delta^2}{2}} 
\le  Te^{-\frac{\delta \Delta^2}{2}},
\end{align*} 
where we again used Hoeffding's inequality and we bound the number
of active intervals by $T$. Thus, by union bound, we get
$$
\probs{\bar{X}\cup\bar{Y}} \le e^{-\frac{\Delta^2 \abss{I}}{2}}
 + T^2e^{-\frac{\delta \Delta^2}{2}}.
$$
Then, by choosing $\delta=\frac{6}{\Delta^2}\log T$, and considering the two cases with $\abss{I} \le \delta$ and $\abss{I} > \delta$, we get the desired result.

\section{Proof of Lemma~\ref{LEM:CASE2}}
The proof follows the same line of reasoning as the previous
lemma. Let $\tilde R_I$ be as in the proof of Lemma~\ref{LEM:CASE1}. 
Let $C > 0$ be a constant to be chosen later and define  
$W = \max\{W_2 - W_1, C\log T\}$.
Let $X$, $Y$ and $E_t$ be the events defined in the proof of Lemma~\ref{LEM:CASE1}
The same argument as before shows that 
\begin{equation*}
\tilde{R}_I \leq \delta + 1 + 2T \probs{\bar X \cup \bar Y}
\end{equation*}
 We proceed to bound $\probs{\bar{X} \cup \bar Y}$. Without loss of generality, we assume that
 $\Delta = \EXPS{r_t(1) - r_t(2)}$. We first consider
 the case where $|I| \ge \frac{2}{\Delta}W$. 

For event $\bar{X}$, we have

\begin{align*}
\probs{\bar{X}} & = \probs{W_q(1, H^1) < W_q(2, H^2)} \\
&= \prob{W_1 + \sum_{t=p}^q r_t(1) <  W_2 + \sum_{t=p}^q r_t(2)} \\
&=\prob{ \left( \sum_{t=p}^q r_t(1) - r_t(2) \right) - |I|\Delta < W_2 - W_1 - |I|\Delta},
\end{align*}
and since $|I| \ge \frac{2}{\Delta}W \ge \frac{2}{\Delta}(W_2 - W_1)$,
by Hoeffding's bound we have 
$$
\probs{\bar{X}} \le \probs{ ( \sum_{t=p}^q r_t(1) - r_t(2) ) - |I|\Delta <
  -\frac{1}{2}|I| \Delta} \le e^{-\frac{\Delta^2|I|}{8}}.
$$
To bound $\probs{\bar Y}$ we again bound the probability of the events
$E_t$ for $t \ge p+ \delta$. Using the exact same technique as in
Lemma~\ref{LEM:CASE1} and letting $\delta = \frac{2}{\Delta}W$ we get 
\begin{equation*}
\probs{E_t} \leq e^{-\frac{\Delta^2 \delta}{8}} 
+ T e^{-\frac{\Delta^2 \delta}{2}}.
\end{equation*}

Then, by union bound, we obtain
$$
\probs{\bar{X}\cup\bar{Y}} \le e^{-\frac{\Delta^2 |I|}{8}} +
Te^{-\frac{\Delta^2 \delta}{8}} + T^2 e^{-\frac{\Delta^2 \delta}{2}}.
$$
Recall that $\delta = \frac{2}{\Delta}W = \frac{2}{\Delta}\max\{W_2 -
W_1, C\log T\}$ and $|I| \ge \frac{2}{\Delta}W = \delta$.  A simple
calculation verifies that setting $C=\frac{8}{\Delta}$, yields
\begin{align*}
\tilde{R}_I \le \delta + O(1) \le \frac{2}{\Delta} (W_2 - W_1 + C\log T) + O(1) 
\le  \frac{2}{\Delta} (W_2-W_1) + \frac{16}{\Delta^2}\log T + O(1).
\end{align*}
On the other hand if $|I| < \frac{2}{\Delta}W$, we have
$$
W_q(1, H^1) - W_1 \le \frac{2}{\Delta}W,~W_q(2, H^2) - W_1 \le \frac{2}{\Delta}W + W_2 - W_1.
$$
Therefore we must have
$$
\max\{W_q(1, H^1), W_q(2, H^2), 0 \} - W_1 \le \frac{2}{\Delta}W + W_2 - W_1,
$$
using the definition of $W$ and bounding the $\max$ by a sum yields
$$
\tilde{R}_I \le \Big(1+\frac{2}{\Delta}\Big) (W_2-W_1) + \frac{16}{\Delta^2}\log T + O(1).
$$

\section{Proof of Lemma~\ref{LEM:CASE3}}
Using the same notation as in the previous two lemmas, we strive to bound
\begin{equation*}
\tilde{R}_I := \EXPS{\max\{W_q(1, H^1), W_q(2, H^2), 0 \} - W_1 }
\end{equation*}
under the assumption that  $W_2 - W_1 > C\log T$ for $C
=\frac{6}{\Delta^2}$.  Let $\xi = \max\{W_q(1, H^1), W_q(2, H^2), 0 \}
- W_1$ and define the sopping time
\begin{align*}
\eta := \inf \{ q \geq t \geq p~|~
\exists J \in \ACT(t) \ \text{with} \ W_t(1, J) > W_t(2, H^2)\},
\end{align*}
with $\eta  = \infty$ if the event does not
occur. Notice that by definition of $\eta$, expert $1$ is chosen
for the first time at time $\eta + 1$.  We consider the
following three disjoint events:
\begin{align*}
  & \cU := \{ \eta = \infty\}  \\
  & \cV := \{ \eta < \infty \wedge W_\eta(1, H^1) \leq  W_\eta(2, H^2) \}\\
  & \cW:=\{\eta < \infty \wedge W_\eta(1, H^1) > W_\eta(2, H^2)\}. 
\end{align*}
We make the following observations:

\begin{enumerate}[(i)]
\item $W_t(2, H^2) = W_2$ for $t \leq \eta$. Indeed, since by definition of
$\eta$ we have $x_t = 2$ for $t \leq \eta$ it follows that $W_t(2, H^2)$
remains constant. 
\item If $\cU$ happens then $\xi = W_2 -W_1$. By definition of $\eta$ we must have $W_q(1, H^1) \le W_q(2, H^2) =
W_2$, which implies the statement. 
\item If $\cW$ happens then we must have 
$$
W_{\eta -1}(1, H^1) \le W_{\eta-1}(2, H^2) = W_2,
$$
and this easily implies
\begin{equation*}
W_2 < W_\eta(1, H^1) \le W_2 + 1.
\end{equation*}
\item $\probs{\eta = l} = 0$ if $ l - (p-1) \le  C \log T $. This is
immediate since $W_1$ is the largest weight over expert $1$ at time
$p-1$ and by assumption $W_2 - W_1 > C\log T$.
\end{enumerate}

Using observation (ii) can write $\tilde R_I$ as follows:
\begin{align}
  \tilde R_I
 & =\EXPS{ \xi   \mid \cU }\probs{\cU} + \EXPS{\xi \mid
               \cV } \probs{\cV} + \EXPS{\xi \ind_\cW} \nonumber \\  
& = (W_2 - W_1) \probs{\cU} +  \EXPS{\xi \mid
               \cV }  \probs{\cV} + \EXPS{\xi \ind_{\cW}}. \label{eq:ridecomp}
\end{align}
We now proceed to bound the last two summands in the above
expression. The main intuition  behind our proof is that since we have
$W_2 - W_1 > C\log T$ it is unlikely that $W_\eta(1, J) > W_\eta(2, H^2)$ for any
interval different from $H^1$.  This will imply $P(\cV)$ is small. On
the other hand if $\cW$ occurs then the algorithm behavior after
$\eta$ is similar to FTL and thus the weights $W_t(1, H^1)$ and $W_t(2, H^2)$ does not
significantly increase for $t > \eta$. 

We proceed to bound $\probs{\cV}$. Let $\bar J = [p_{\bar J}, q_{\bar
  J}] \in \I$ be such that $W_\eta(1, \bar J) > W_\eta (2, H_2)$. Using the same argument of observation (iv) we
must have $|\bar J| \geq C \log T$. Furthermore, by definition of $\cV$,
we must have $p_{\bar J}  > p$ (again, we use $p_J$ to denote the starting time of interval $J\in\I$). Indeed, if $\bar J$ starts at or before $p$, then by
defintion of $H^1$, $W_{p-1}(1, H^1) \ge  W_{p-1}(1, \bar J)$
which would imply $W_{\eta}(1, H^1) \ge W_\eta(1,\bar  J)$
contradicting the definition of $\cV$. Finally, by construction of
$\I$ if $\bar J$ does not start at $p$ then we must have $p_{\bar J} - p \ge |\bar J|
\geq C \log T$.  Therefore we have:
\begin{align*}
  \probs{\cV} & \leq  \probs{W_{\eta}(1, H_1) < W_\eta(1,\bar J)} \\
&  =  \prob{W_1 + \sum_{t=p}^{\eta} r(1) - r(2) < \sum_{t=p_{\bar J}}^{\eta}
  r(1) - r(2) } \\
& = \prob{ W_1 + \sum_{t=p}^{p_{\bar J} - 1}r(1) - r(2) < 0} \\
& \leq \prob{ \sum_{t=p}^{p_{\bar J} - 1} r(1) - r(2) < 0}. 
\end{align*}
We can use the union bound and the observation that $p_{\bar J} - p \ge
C \log T$ to bound the above probability as
\begin{align*}
 \prob{ \sum_{t=p}^{p_{\bar J} - 1} r(1) - r(2) < 0} 
 \leq  \sum_{\substack{J \in \I \\  p_J - p \ge C \log T}} \prob{\sum_{t=p}^{p_J - 1}
  r(1) - r(2) < 0} 
\leq  2 T e^{-\frac{\Delta^2 (p_J - p)}{2}}. 
\end{align*}
where applied Hoeffding's inequality, the fact that $\E[r(1) - r(2)] \ge
\Delta$ and the fact that $| \I | \leq 2 T$.  Finally, using the fact
that $p_J - p \ge C \log T $  and replacing the value of $C$ we see that
\begin{equation}
\label{eq:pv}
\probs{V} \leq \frac{2}{T^2}
\end{equation}
We proceed to bound $\EXPS{\xi \ind_{\cW}}$. Notice that from
observation (iii) we have 
\begin{align*}
  \xi = & \max\{W_q(1, H^1) , W_q(2, H^2) , 0\} - W_2 + W_2 - W_1  \\
\leq & \max\{W_q(1, H^1), W_q(2, H^2) , 0\} - W_\eta(1, H^1) + 1 + W_2 -W_1.
\end{align*}
Let $\xi_\eta = \max\{W_q(1, H^1), W_q(2, H^2) , 0\} - W_\eta(1, H^1)$.
  Fix $\delta > 0$ and  notice that if  $q
- \eta \ge \delta$ we necessarily most have $\xi_\eta \le \delta$. Therefore,
we focus on the  case where $q - \eta > \delta$.
As in the two previous lemmas  define $\tau = \inf\{q >  t \geq
\eta~:~W_t(1, H^1) \geq W_\eta(1, H^1) + \delta\}$ and $\tau = \infty$ if $ W_t(1,
H^1) < W_\eta(1, H^1) + \delta$ for all $t \in [p, q]$. Define also the events:
$$ 
X:=\{W_q(1, H^1) \ge W_q(2, H^2)\},
$$
$$
Y:=\{\tau < q \ \wedge \ x_t = 1 \ \forall\, t \in [\tau + 1, q]  \}
\cup \{\tau = \infty\}.
$$
With the same arguments as the previous two lemmas we can show that if $X\cap Y$
happens we most have $\xi_\eta < \delta + 1$. It follows that:
\begin{equation}\label{eq:xiw}
\begin{aligned}
\EXPS{\xi \ind_{\cW}}
\leq  (\delta + 2 + W_2 - W_1) \probs{\cW} 
+ \EXPS{\xi_\eta + 1 + W_2 - W_1 \mid \bar X \cup \bar Y, \cW} \probs{\bar X \cup \bar Y, \cW}. 
\end{aligned}
\end{equation}
Notice that if $W_q(1, H_1) < W_q(2, H_2)$ and $\cW$ happens we must
have $\sum_{t= \eta+1}^q r_t(1) - r_t(2) < 0$. Using the tower property of
conditional expectation we have 
\begin{align*}
\probs{\bar X, \cW}  =  \EXPS{\probs{\bar X, \cW | \eta}} 
 \leq \EXP{ \prob{\sum_{t=\eta+1}^q r_t(1) - r_t(2) < 0| \eta}}
 \leq \EXPS{e^{-\frac{\Delta^2(q - \eta)}{2}} } \leq e^{-\frac{\Delta^2 \delta}{2}}.
\end{align*}
Here for the last inequality we use Hoeffding's inequality and the fact that $q - \eta >
\delta$ by assumption. Notice that, here, although we condition on $\eta$, i.e.,
the event $W_\eta(1, H^1) > W_\eta(2, H^2)$, this event only depends on the
random rewards up to time $\eta$, and does not affect the independence of rewards starting
from $\eta+1$, and therefore we can apply Hoeffding's inequality.
To bound the probability of $\bar Y \cup \cW$
we again introduce the events
$$
E_t := \{t > \eta , W_{t-1}(1, H^1)> W_\eta(1, H^1) + \delta~\wedge~ x_t=2\},
$$
And notice that $\bar Y = \bigcup E_t$. Moreover notice that if $ t -
\eta  \le \delta$ then $\probs{E_t | \eta} = 0$. Therefore we can assume
that $t - \eta > \delta$.  Define $I_t$ as in~\eqref{eq:def_it}.  We can now bound $\probs{E_t ,
\cW}$ as follows:
\begin{equation}\label{eq:condsumdecomp}
\begin{aligned}
\probs{E_t , \cW} =&\EXPS{\probs{E_t , \cW | \eta}} \\
=& \mathbb{E}[\mathbb{P}\{W_{t-1}(1, H^1) > W_\eta(1, H^1) + \delta,x_t = 2, \cW | \eta\}] \\
\leq & \EXPS{\probs{W_{t-1}(1, H^1) < W_{t-1}(1, H^2), \cW | \eta}}  \\
&+\sum_{\substack{J\in\ACT(t) \\ p_J>p}}
 \EXPS{\probs{W_{t-1}(1, H^1) > \delta,
  I_t= J, x_t = 2 , \cW \mid \eta}}.
\end{aligned} 
\end{equation}

Notice that since we condition on $\eta$, we have $W_{\eta}(1, H^1) > W_\eta(2,
H^2)$. Therefore the even in the first summand of the above expression
happens if and only if $\sum_{s=\eta+1}^{t-1} r_s(1) - r_s(2)\leq
0$. Using Hoeffding's bound we can thus bound the first term by 
\begin{equation}
\label{eq:hoeffdingfirst}
\begin{aligned}
   \EXPS{\probs{W_{t-1}(1, H^1) < W_{t-1}(1, H^2), \cW | \eta}} 
\leq \EXPS{e^{-\frac{\Delta^2(t - \eta)}{2}}} \leq e^{-\frac{\Delta^2 \delta}{2}}.
\end{aligned}
\end{equation}
We proceed to bound the second term in
\eqref{eq:condsumdecomp}. We split the sum to cases where $J = [p_J,
q_J]$ is such that $p_J \le \eta$ and that $p_J > \eta$. For the first
case, notice that since $W_t(2, H^2)$ remains unchanged for $t=p,\ldots, \eta$,
$x_t=2$ for all $t=p,\ldots, \eta$, and $p_J > p$, we have $W_\eta(2, H^2) > W_\eta(2, J)=0$.
Therefore, for all $t>\eta$, $W_t(2, H^2) > W_t(2, J)$, and thus $I_t=J$ cannot happen.

For the second case, if $p_J  > \eta$ notice that the event can only
happen if $W_{t-1}(2, J) > \delta$ and $W_{t-1}(2,
J) > W_{t-1}(1, J) $. This implies that $t - p_J > \delta$ and the same
argument as before shows  
\begin{equation}
\label{eq:hoeffdinglast}
\probs{W_{t-1}(1, H^1) > \delta,
  I_t= J, x_t = 2 , \cW \mid \eta} \leq e^{-\frac{\delta
    \Delta^2}{2}}.
\end{equation}
Using inequalities \eqref{eq:hoeffdingfirst} and \eqref{eq:hoeffdinglast} and the fact that
there are at most $T$ active intervals at time $t$ we see that
\eqref{eq:condsumdecomp} can be bounded as:
\begin{equation*}
  \probs{E_t , \cW}  \leq T e^{-\frac{\delta \Delta^2}{2}}.
\end{equation*}
By the union bound we thus have
\begin{equation*}
  P(\bar X \cup \bar Y , \cW) \leq (1 + T^2)  e^{-\frac{\Delta^2 \delta
    }{2}} \leq 2 T^2 e^{-\frac{\Delta^2 \delta
    }{2}}.  
\end{equation*}
Replacing this in \eqref{eq:xiw} we see that:
\begin{equation*}
\EXPS{\xi \ind_{\cW}} \leq  (\delta  + 2  T^3
e^{-\frac{\Delta^2\delta}{2}}  + W_2 - W_1 + 2) \probs{\cW}
\end{equation*}
Finally, letting $\delta = \frac{6}{\Delta^2} \log T$ we obtain:
\begin{equation}
\EXPS{\xi \ind_{\cW}} \leq \Big( \frac{6}{\Delta^2} \log T + W_2 - W_1 +
O(1) \Big) \probs{\cW}. 
\end{equation}
Finally combining this bound with \eqref{eq:ridecomp} and \eqref{eq:pv}
we obtain the desired bound.